\documentclass[10pt,twocolumn,letterpaper]{article}

\usepackage{titling}
\usepackage{iccv}
\usepackage{bm}
\usepackage{times}
\usepackage{epsfig}
\usepackage{graphicx}
\usepackage{amsmath}
\usepackage{amssymb}
\usepackage{pifont}
\usepackage{float}
\usepackage{multirow}
\usepackage{subfig}
\usepackage{graphics}
\usepackage{adjustbox}
\usepackage{multicol}
\usepackage{url}
\usepackage{enumitem}
\usepackage{authblk}
\usepackage{makecell}

\usepackage[pagebackref=true,breaklinks=true,colorlinks,bookmarks=false]{hyperref}

\makeatletter
\renewcommand\AB@affilsepx{, \protect\Affilfont}
\makeatother

\makeatletter
\def\blfootnote{\xdef\@thefnmark{}\@footnotetext}
\makeatother

\DeclareMathOperator*{\argmin}{argmin}
\newcommand{\boldparagraph}[1]{\vspace{0.3em}\noindent{\bf #1} }
\newcommand{\cmark}{\ding{51}}%
\newcommand{\norm}[1]{\left\lVert#1\right\rVert}

\iccvfinalcopy %
\def\iccvPaperID{6178} %

\ificcvfinal\fi %

\begin{document}

\title{\vspace{-1cm}H2O: Two Hands Manipulating Objects for First Person Interaction Recognition \vspace{-0.4cm}}

\author{Taein Kwon$^1$, Bugra Tekin$^2$, Jan St{\"u}hmer$^3$\thanks{Work performed while
		at Microsoft.} , Federica Bogo$^1$, and Marc Pollefeys$^{1,2}$\\
	$^1$ETH Zurich, $^2$Microsoft, $^3$Samsung AI Center, Cambridge
}
\maketitle
\thispagestyle{empty}
\begin{abstract}
	We present a comprehensive framework for egocentric interaction recognition using markerless 3D annotations of two hands manipulating objects.
	To this end, we propose a method to create a unified dataset for egocentric 3D interaction recognition. Our method produces annotations of the 3D pose of two hands and the 6D pose of the manipulated objects, along with their interaction labels for each frame. Our dataset, called H2O (2 Hands and Objects), provides synchronized multi-view RGB-D images, interaction labels, object classes, ground-truth 3D poses for left \& right hands, 6D object poses, ground-truth camera poses, object meshes and scene point clouds. To the best of our knowledge, this is the first benchmark that enables the study of first-person actions with the use of the pose of both left and right hands manipulating objects and presents an unprecedented level of detail for egocentric 3D interaction recognition.
	We further propose the method to predict interaction classes by estimating the 3D pose of two hands and the 6D pose of the manipulated objects, jointly from RGB images. Our method models both inter- and intra-dependencies between both hands and objects by learning the topology of a graph convolutional network that predicts interactions. We show that our method facilitated by this dataset establishes a strong baseline for joint hand-object pose estimation and achieves state-of-the-art accuracy for first person interaction recognition.
\end{abstract} \blfootnote{Project Page: \url{https://www.taeinkwon.com/projects/h2o}}

\vspace{-3mm}
\section{Introduction}\label{sec:introduction}
\begin{figure}[t]
\begin{center}
   \includegraphics[width=1.0\linewidth]{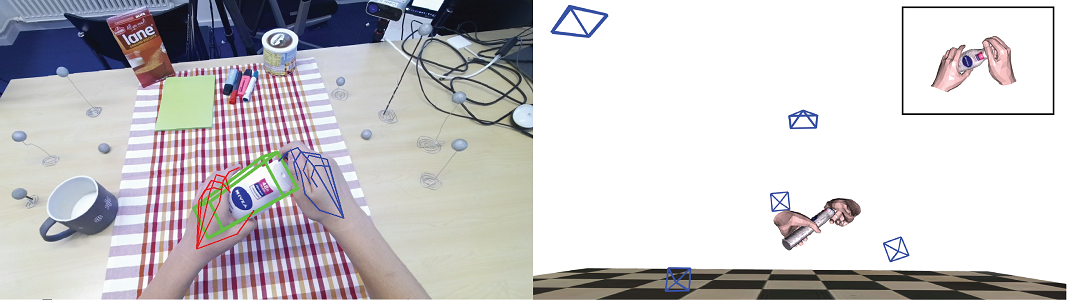}
\end{center}
\vspace{-5mm}
   \caption{\small {\bf Two hands manipulating objects for first person interaction recognition.} We propose a dataset providing rich annotations for 3D poses of left \& right hands, 6D object poses, camera poses, object meshes and scene point clouds, along with their associated interaction labels. We leverage our dataset to propose novel methods for 3D interaction recognition.  %
   }
\label{fig:teaser}
\vspace{-5mm}
\end{figure}

In recent years, there has been tremendous progress in video understanding and action recognition. Current algorithms can reliably recognize the action the subject is performing in many unconstrained settings from third person viewpoints~\cite{carreira2017quo,feichtenhofer2020x3d,feichtenhofer2019slowfast,feichtenhofer2016convolutional,simonyan2014two,wu2019long}. Although action recognition from first-person views has many applications in augmented reality, robotics and surveillance, it trails behind the progress in third person views, mostly due to the lack of large and diverse egocentric datasets. From an egocentric viewpoint, action recognition is mostly about understanding hand \& object interactions. A unified understanding of the positions and movements of hands and the manipulated objects is crucial for recognizing egocentric interactions. However, existing first-person interaction datasets mostly provide only 2D features (\eg bounding boxes, hand segmentation) without reasoning in 3D about the motions of hands and the manipulated objects. In this work, we propose, for the first time, a unified dataset for first person interaction recognition with markerless 3D annotations of two hands manipulating objects, as depicted in Fig.~\ref{fig:teaser}. We collect a richly annotated %
dataset 
including synchronized \mbox{RGB-D} images, camera poses, right \& left hand poses, object poses, object meshes, scene point clouds and action labels, which provides an unprecedented level of detail for understanding 3D hand-object interactions. With the help of our dataset, we  present the first method to estimate jointly the 3D pose of two hands and objects from a color image. We further propose to learn  interdependencies within and across hand and object poses using an adaptive graph convolutional network for 3D interaction recognition.

Jointly capturing hands in action and the manipulated objects in 3D is a challenging problem due to reciprocal occlusions. The problem is more challenging from first person viewpoints due to the unique challenges brought by egocentric vision such as fast camera motion, large occlusion, background clutter~\cite{li2015delving} and most importantly, lack of datasets. Recent works have proposed datasets that successfully addressed some of these challenges. Sridhar et al.~\cite{sridhar2016} have presented one of the earliest datasets for hand-object interactions, in which a single hand manipulates a cuboid object. Pioneering works by~\cite{garcia2018first,hampali2020honnotate,hasson2019learning} have further proposed datasets that include 3D annotations for object manipulation scenarios of a single hand.

Most of these works, however, are limited by different factors. They mainly focus on \emph{single hand manipulation scenarios}~\cite{garcia2018first,hampali2020honnotate,hasson2019learning}. While single hand manipulation is relevant for some scenarios, most of the time, hand-object interaction involves two hands manipulating an object. Using \emph{only 2D annotations},~\cite{tzionas2016,tzionas2014} presented datasets for hand-hand and hand-object interactions. The intricate nature of hand-object interactions, however, requires 3D reasoning rather than 2D to better resolve mutual occlusions. In the context of hand-object interactions, early work mostly tackles the problem of joint estimation of 3D hand and object poses, \emph{without reasoning about the actions}. While precise 3D position data for hands and objects is crucial for many applications in robotics and graphics, the sole knowledge of the pose lacks semantic meaning about the actions of the subject. To that end,~\cite{garcia2018first} released an egocentric action dataset including 3D annotations of hands and objects; however, the data is captured with an \emph{intrusive motion capture system}. Although motion capture datasets~\cite{garcia2018first,taheri2020grab} can provide large amounts of training samples with accurate 3D annotations, they can only be captured in controlled settings and have visible markers on the images that bias pose prediction in color images. \emph{Synthetic datasets}~\cite{hasson2019learning} could provide an alternative to them, however, the existing ones cannot yet reach the realism that is needed to generalize to real images and are only for single-image scenarios that lack temporal context crucial for recognizing interactions. 

Our method aims at tackling these limitations exhibited by prior work. To this end, we propose an approach for creating a unified dataset for egocentric 3D interaction recognition that includes markerless annotations of the 3D pose of two hands and the 6D pose of the manipulated objects, along with their associated action labels for each frame of a large number of recordings that include 571,645 synchronized \mbox{RGB-D} frames. In addition, we propose the first method to jointly predict the 3D pose of two hands and 6D pose of the manipulated objects using only RGB images and present a novel 3D interaction recognition approach that learns the interdependencies between hand and object poses by a topology-aware graph convolutional network.

Our contributions can be listed as follows:
\vspace{-3mm}
\begin{itemize}[leftmargin=12pt]
    \item We present the first unified dataset for egocentric interaction recognition with markerless 3D annotations of two hands and the 6D pose of manipulated objects. Our dataset, which we call \emph{$H2O$}, standing for \emph{2 hands} and \emph{objects}, provides rich ground-truth annotations for 3D hand-object poses \& shapes, action labels, camera poses, scene point clouds and object meshes that enable us to produce comprehensive egocentric scene interpretations.
    \vspace{-6mm}
    \item We propose a semi-automatic pipeline to curate a hand-object interaction dataset with action labels and the poses of two interacting hands as well as the objects in contact, using a practical multi-camera system with diverse backgrounds. We demonstrate the fidelity and accuracy of our annotations by detailed verifications.
    \vspace{-2mm}
    \item We introduce a unified approach to recognize hand-object interactions from RGB images that simultaneously predicts, for the first time, the 3D pose of two interacting hands and the 6D pose of manipulated objects, along with action and object classes. 
    \vspace{-2mm}
    \item Leveraging our dataset, we propose a novel method for 3D interaction recognition that learns the interdependencies between two hands and objects with a topology-aware graph convolutional network. To this end, we parameterize both hand and object poses as individual graphs and combine them in a single multi-graph architecture. We then learn the interdependencies and connections between different graph entities with an adaptive architecture and compute the topology of the multi-graph structure for recognizing 3D hand-object interactions. 
    \vspace{-2mm}
\end{itemize}

We demonstrate that using the pose predictions facilitated by our dataset, we achieve better overall performance for recognizing interactions outperforming the state-of-the-art~\cite{carreira2017quo,fan2020pyslowfast,feichtenhofer2019slowfast}. We further provide baselines for hand \& object pose estimation and interaction recognition to enable further benchmarking on this dataset. We will make our dataset and annotations publicly available upon acceptance.

\begin{figure*}[]
\begin{center}
\vspace{-4.5mm}
\includegraphics[width=1.0\linewidth]{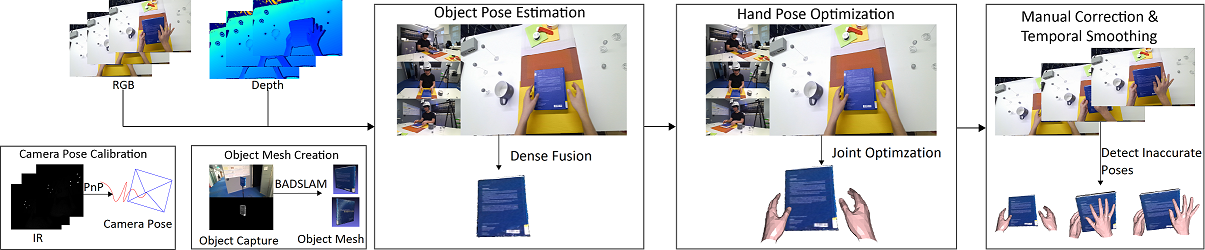}
\end{center}
\vspace{-5mm}
\hspace{1.25cm}(a)\hspace{2.2cm}(b)\hspace{2.8cm}(c)\hspace{3.8cm}(d)\hspace{3.8cm}(e)
\vspace{-3mm}
\caption{\small (a) We calibrate cameras using IR sphere markers and PnP~\cite{lepetit09}, (b) create object meshes using BADSLAM~\cite{schops2019bad} on RGB-D captures, and (c) estimate object poses using DenseFusion~\cite{wang2019densefusion} on \mbox{RGB-D} images and mask images from Mask R-CNN \cite{he2017maskrcnn}. We then select the pose with the highest confidence among five cameras. (d) Consequently, we detect hand joints with OpenPose~\cite{cao2018openpose} and optimize hand shape using Eq.~\ref{eq:loss_sum}. (e) We finally detect and smooth temporally inaccurate poses.}
\label{fig:pipeline}
\vspace{-5mm}
\end{figure*}

\vspace{-1mm}
\section{Related Work}\label{sec:related_work}
\vspace{-1mm}

\boldparagraph{Datasets for egocentric action recognition and hand-object pose estimation.} While many datasets for third-person action recognition have been proposed throughout the years~\cite{gu2018ava,kay2017kinetics,sigurdsson2016Charades,zhou2018towards}, recently there is a surge in interest for data targeting also egocentric scenarios~\cite{damen2020epickitchens,Damen2018EPICKITCHENS,goyal2017something, li2018eye,ragusa2020meccano,sigurdsson2018charades} that involve mostly 2D features. 
These datasets provide only limited multi-view data and do not provide hand and object poses, which have been shown to be useful cues for a comprehensive understanding of the scene~\cite{garcia2018first,tekin2019h+}.

\begin{table}[b]
\vspace{-3mm}
\begin{center}
\begin{adjustbox}{width=1.1\columnwidth,center}
\begin{tabular}{l|c |c| c| c| c| c| c| c| c| c}
\hline
Dataset & Frames & Action & 6D Obj & 3D left & 3D right & (**)Markerless & Real & Ego & Depth  & Multiview\\
\hline
H2O & 571k & \cmark & \cmark & \cmark & \cmark & \cmark & \cmark & \cmark & \cmark & \cmark \\
\hline
FPHA \cite{garcia2018first} & 100k & \cmark & * (23k) & $\cdot$ & \cmark & $\cdot$ & \cmark & \cmark & \cmark  & $\cdot$ \\

HOnnotate \cite{hampali2020honnotate} & 78k & $\cdot$ & \cmark & $\cdot$  & \cmark & \cmark & \cmark & $\cdot$  & \cmark   & \cmark \\
Obman \cite{hasson2019learning} & 150k & $\cdot$ & \cmark & $\cdot$ & \cmark & \cmark &$\cdot$   &$\cdot$   & \cmark  & $\cdot$ \\
Freihand \cite{zimmermann2019freihand} & 37k & $\cdot$ & $\cdot$  & $\cdot$ & \cmark & \cmark & \cmark & $\cdot$  & $\cdot$    & \cmark \\
Panoptic \cite{Joo2017panoptic} & 1.5M & $\cdot$ & $\cdot$  & \cmark & \cmark & \cmark & \cmark & $\cdot$  & \cmark    & \cmark \\
ContactPose \cite{brahmbhatt2020contactpose} & 2.9M & $\cdot$ & (***) \cmark  & \cmark & \cmark & $\cdot$ & \cmark & $\cdot$  &  \cmark     & \cmark \\ %
\hline
\end{tabular}
\end{adjustbox}
\end{center}
\vspace{-5mm}
\caption{\small A comparison of the existing related \emph{image-based datasets} with 3D annotations for hand interactions. H2O provides a total of 571k frames captured from 5 different views.
(*): Object pose provided only for a subset of frames.  (**) Methods without markers on hands \& objects. (***) Printed, textureless objects.}
\label{table:other_datasets}
\vspace{-5mm}
\end{table}

A few datasets collect hand pose ground truth, acquired in an automated or semi-automated way (Panoptic~\cite{joo2015panopticstudio,Joo2017panoptic}, FreiHand~\cite{zimmermann2019freihand}, Interhand~\cite{moon2020interhand2}). However, they do not consider interactions with objects.
Recently, GRAB~\cite{taheri2020grab} uses a mocap system and objects from~\cite{brahmbhatt2019contactdb} to track body and hand pose while interacting with the scene without providing corresponding images. 
HOnnotate~\cite{hampali2020honnotate} relies on an optimization process to estimate accurate hand and object pose from multi-view RGB-D data. ObMan~\cite{hasson2019learning} collects purely synthetic images of hands holding objects. All these works, however, consider only single-hand scenarios and do not focus on action recognition. Similarly to us, FPHA~\cite{garcia2018first} collects egocentric RGB-D frames with action, hand and object pose annotations. However, the dataset relies on mangnetic sensors, which pollute the RGB images, and does not include neither multi-view data nor two-hand poses.

As shown in Table~\ref{table:other_datasets}, our dataset is the first including real, multi-view RGB-D data and accurate annotations for the 3D pose of two hands, object pose, and action labels for egocentric 3D interaction recognition.

\boldparagraph{Hand \& object pose estimation.} While a significant amount of research has focused on predicting the pose of hands~\cite{ge2018hand, moon2018v2v-posenet,mueller2017real,oberwger2017deepprior++,simon2017hand,ye2016spatial,yuan2018depth,Zimmermann2017learning} or objects~\cite{brachmann2016uncertainty,li2018deepim,peng2019pvnet,tekin2018real-time,wang2019densefusion,xiang2017posecnn} in isolation, joint understanding of hand-object interactions has received far less attention.
Considering hands and objects together adds a number of challenges, which require to reason \eg about occlusions and inter-penetrations.
Pioneer works in~\cite{ballan2012motion,tzionas2016,tzionas2014} investigate hand-hand and hand-object interactions relying on optimization frameworks which might be slow and difficult to tune. Tekin et al.~\cite{tekin2019h+} and Hasson et al.~\cite{hasson2020leveraging,hasson2019learning} efficiently estimate hand \& object poses directly from RGB images. %
However they consider only single-hand scenarios.

\boldparagraph{Recognizing interactions.} Action recognition has received a lot of attention in the computer vision community~\cite{bobick2001recognition,dollar2005behavior,jhuang2007biologically,laptev2005space,niebles2008unsupervised,wang2011action,wong2007learning}. With the advent of deep learning and the availability of large datasets, significant progress has been made in third-person action recognition~\cite{carreira2017quo,feichtenhofer2020x3d,feichtenhofer2019slowfast,feichtenhofer2016convolutional,lin2019tsm,wu2019long,wang2016temporal,zhou2018temporal}. Recently, there has also been an increase in interest for explicitly reasoning about human-object interactions~\cite{escorcia2013spatio,fang2018pairwise,gkioxari2018detecting,gupta2019no,kato2018compositional,li2019transferable,rosenfeld2016hand,wan2019pose,wang2019deep,xiao2019reasoning,zhou2019relation} and skeletal action recognition~\cite{cheng2020skeleton,li2019actional,liu2020disentangling,shi2019two,yan2018spatial}, however mostly from third-person viewpoints.

Recognizing interactions from first-person viewpoints, however, poses a number of specific challenges like large occlusions, fast camera motion and background clutter~\cite{li2015delving}. While initially the lack of large amounts of data somewhat hindered the development of effective DNN-based methods, over the recent years, there has been a renewed interest in the problem.
Some approaches leverage multi-modal input like head motion~\cite{kitani2011fast, li2013learning, ryoo2013first,singh2016first} and eye gaze~\cite{fathi2012learning, li2013learning}. It is also common to extract features with CNNs and leverage additional 2D cues related to motion, hand location, object location or object class, in isolation~\cite{bertasius2016first,li2013learning,ma2016going} or jointly ~\cite{fathi2011understanding,fathi2011learning,fouhey2018lifestyle,sigurdsson2018actor,sundaram2009high}. While all these methods focus on 2D features, recent work~\cite{garcia2018first,tekin2019h+} suggests that 3D cues (like hand and object pose) can be effective in the context of egocentric action recognition.
However, existing methods have focused on single-hand tracking and no attention has been paid so far to estimating the pose of two hands interacting with objects -- a scenario which is more representative of the interactions encountered in real-world scenarios.

\vspace{-2mm}
\section{Annotation Method}\label{sec:method}
\vspace{-1mm}

Fig.~\ref{fig:pipeline} shows an overview of our annotation pipeline.
We capture synchronized RGB-D frames from multiple views with five Azure Kinect cameras~\cite{azurekinect2020}.
One of the cameras is mounted on a helmet worn by different subjects to capture egocentric frames. We acquire ground-truth hand and object poses in a semi-automated way.
First, we scan each object with a Kinect to obtain a complete 3D model. This model is used to track object 6D pose in each frame via DenseFusion~\cite{wang2019densefusion}. To track hands, we fit the MANO parametric hand model~\cite{romero2017embodied} to multi-view depth data in each frame. This automated tracking process may fail on some frames, due to challenges like \mbox{(self-)occlusions}, blur and cluttered background. We therefore manually detect failure cases and remove corresponding poses; such poses can be then replaced via temporal smoothing.
Finally, we manually annotate action labels over the sequences.
In the following sections, we describe each step of our pipeline in detail.

\vspace{-1mm}
\subsection{Camera Calibration}\label{sec:camera_pose_estimation}
\vspace{-1mm}
Our setup consists of four static plus one head-mounted RGB-D cameras. We use the factory-calibrated intrinsic parameters accessible via Azure Kinect DK~\cite{azurekinect2020}. As for the extrinsic parameters, we obtain them with a calibration method relying on IR reflective spheres. We choose this method to make our setup portable and easy to deploy.

We place nine IR reflective spheres at random locations in the scene, ensuring that each sphere is visible from all the cameras. In the IR images captured by our cameras to reconstruct depth, such spheres are shown as bright circles, which can be easily detected in an automated way. We compute the center of each sphere and then obtain its 3D location by considering the corresponding pixel in the depth image. Given the 3D location of the nine spheres in each frame, we solve for camera pose via PnP~\cite{lepetit2009epnp}. In order to consistently identify spheres across frames, we define an initial mapping in the first frame and then track it over time.

Poses computed for the head-mounted camera can exhibit jitter. We smooth them via Kalman filtering~\cite{wan2000unscented}, under the assumption that the head moves with uniform speed. The overall framework allows us to use multiple cameras during annotation, which eventually increases the fidelity and accuracy of our annotations.

\subsection{Object Pose Annotation}\label{sec:object_pose_estimation}
We obtain accurate per-frame object 6D poses using multi-view images together with camera pose information. We first reconstruct a 3D mesh model for each object. To this end, we scan the object by capturing RGB-D frames with a hand-held Kinect camera moving around it. We feed these frames into a state-of-the-art RGB-D SLAM method, BADSLAM~\cite{schops2019bad}, to reconstruct a 3D mesh.
We obtain texture for each object in Blender~\cite{blender2018}: we project the RGB images obtained at scanning time onto the mesh surface, using the camera pose returned by BADSLAM.

We leverage these models to train an object pose tracker.
First, we train an object mask predictor based on Mask R-CNN~\cite{he2017maskrcnn}. As training data, we use the masks obtained by projecting our 3D models onto the images used for their BADSLAM-based reconstruction. %
Then, we feed mask predictions together with the corresponding RGB-D images into DenseFusion~\cite{wang2019densefusion} to estimate object pose. We obtain a pose prediction for each camera view, and select the one with the highest confidence.
Finally, we refine this pose estimate via ICP~\cite{lepetit2009epnp}. Namely, we compute a point cloud from each of the five depth images and merge them into a single point cloud by using camera pose information; then, we fit our object model to this point cloud, taking the prediction from DenseFusion as initialization.

\subsection{Hand Pose Annotation}\label{sec:hand_pose_estimation}
For hand pose estimation, we rely on the widely used MANO hand model~\cite{romero2017embodied}.
MANO factorizes human hand shape into a set of identity parameters~$\beta \in \mathbb{R}^{10}$ and a set of pose parameters~$\theta \in \mathbb{R}^{51}$, storing angles for 15 skeleton joints plus global rotation and translation.
Formally, we can define MANO as a function $H_V(\theta, \beta)$ returning a triangulated mesh with $N_V$ vertices. %
We also define the MANO skeleton as a function $H_J(\theta, \beta)$ which returns \mbox{$N_J=21$} joint locations (the 15 original ones plus 6 other for fingertips and wrist, to map to the OpenPose~\cite{cao2018openpose} skeleton -- see Supp.~Mat.).
We take the object pose estimated as above, and we leverage it when tracking hand pose.

We track hands by minimizing at each frame, $f$, a loss function defined as:
\vspace{-2mm}
\begin{equation}\label{eq:loss_sum}
  \begin{aligned}
    \hat \theta_{f} = \argmin_\theta\sum_{c=1}^{N_{C}}(\lambda_{1}\mathcal{L}_{s}+\lambda_{2}\mathcal{L}_{2D})+\lambda_{3}\mathcal{L}_{3D}+ \\ \lambda_{4}\mathcal{L}_{p}+\lambda_{5}\mathcal{L}_{phy}+\lambda_{6}\mathcal{L}_{a}+\lambda_{7}\mathcal{L}_{m}
  \end{aligned}
  \vspace{-2mm}
\end{equation}
where $N_{c}$ is the number of cameras. Here, $\mathcal{L}_{s}$ is a silhouette-based error term, $\mathcal{L}_{2D}$ and $\mathcal{L}_{3D}$ measure joint error in 2D and 3D, respectively, $\mathcal{L}_{p}$ and $\mathcal{L}_{a}$ are regularizers for pose, $\mathcal{L}_{phy}$ penalizes physically implausible inter-penetrations between hand and objects, and $\mathcal{L}_{m}$ penalizes distance in 3D between the hand depth data and the MANO surface. Lambdas weight the contribution of each error term.
Note that, in order to obtain subject-specific parameters, we minimize Eq.~(\ref{eq:loss_sum}) with respect to $\bm{\beta}$ on one frame only. Then, we track hand pose over the sequence by keeping $\beta$ fixed and optimizing Eq.~(\ref{eq:loss_sum}) with respect to $\theta$ (for both left and right hands). We omit $\beta$ from the following equations for simplicity.

\boldparagraph{2D joint error.} We penalize distance in 2D between MANO joints and OpenPose estimates by defining:
\vspace{-2mm}
\begin{equation}\label{eq:loss_2d}
  \begin{aligned}
    \mathcal{L}_{2D}(\theta)=\sum_{i=1}^{N_{J}}\norm{J_{2D,c}[i]-\Pi_{c}(H_{J}(\theta)[i])}
  \end{aligned}
  \vspace{-2mm}
\end{equation}
where $J_{2D}$ denotes the 2D joint positions pre-computed with OpenPose,
and $H_J(\theta)[i]$ returns the $i$th 3D joint location of the MANO skeleton.

\boldparagraph{3D joint error.} Similarly to the 2D joint error, we compute a penalty in 3D
by triangulating OpenPose estimates. We found that using this error term helps achieve faster convergence and increase stability.

\boldparagraph{3D mesh surface error.}
We obtain a point cloud for hand data by merging the point clouds obtained from each depth image across our different views, and segmenting out the points that do not project onto the hand mask computed as above. Our 3D surface error term penalizes the distance between this point cloud and the MANO surface:
\vspace{-2mm}
\begin{equation}\label{eq:mesh_error}
  \begin{aligned}
    \mathcal{L}_{m}(\theta)=
    \sum_{i=1}^{N_{V}}\norm{(p_j-H_{V}(\theta)[i])\cdot H^\bot_{V}(\theta)[i]} \\
    \text{where } j = \argmin_{j}\norm{p_j-H_{V}(\theta)[i]}
  \end{aligned}
  \vspace{-2mm}
\end{equation}

\noindent where $p_j$ is the $j$th point of the point cloud, and $H^\bot_{V}(\theta)$ denotes the normal of the hand mesh vertex $i$.

As shown in Eq.~\ref{eq:loss_sum}, our optimization function further includes a silhouette error term and regularizers for joint angle limits and physical constraints. We refer the reader to the Supp. Mat. for more details on these terms and for an ablation study on the influence of different error terms in the annotation accuracy.

After running our automated pipeline, we inspect all the frames to identify and remove inaccurate poses.  As a final step, we smooth and interpolate poses via Kalman filtering.

\begin{figure}[t]
  \begin{center}
    \captionsetup[subfigure]{labelformat=empty}
    \subfloat[Read]{\includegraphics[width=0.113\textwidth]{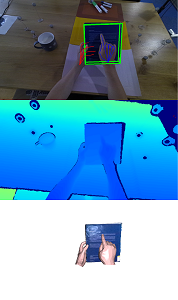}}
    \subfloat[Grab]{\includegraphics[width=0.115\textwidth]{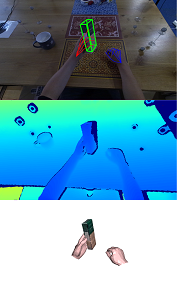}}
    \subfloat[Squeeze]{\includegraphics[width=0.115\textwidth]{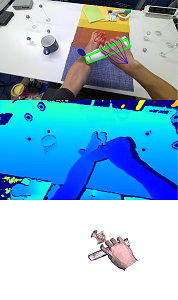}}
    \subfloat[Spray]{\includegraphics[width=0.115\textwidth]{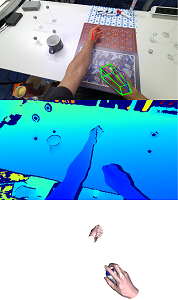}}
  \end{center}
  \vspace{-7mm}
  \caption{\small RGB and depth images with the corresponding annotations of hand \& object pose, and action label. First row: Left hand keypoints, right hand keypoints, and 3D object bounding box are projected on the RGB image. Second row: Synchronous depth images. Third row: Ground-truth data for hand and object meshes. We provide further examples of ground-truth data in Supp. Mat.}
  \label{fig:ground_truth}
  \vspace{-6mm}
\end{figure}

\vspace{-1mm}

\subsection{Temporal Action Annotation}\label{temporal_action_annotataion}
We provide action labels as verb-noun pairs.
We consider 11 verb classes: \emph{grab}, \emph{place}, \emph{open}, \emph{close}, \emph{pour}, \emph{take out}, \emph{put in}, \emph{apply}, \emph{read}, \emph{spray} and \emph{squeeze}. As for nouns, we consider 8 classes: \emph{book}, \emph{espresso}, \emph{lotion}, \emph{spray}, \emph{milk}, \emph{cocoa}, \emph{chips}, \emph{cappuccino}.
By combining verbs and nouns, after excluding pairs which are not represented in our dataset, we obtain a total of 36 action classes.
Note that we pick only one verb and one noun for every frame, so there are no overlapping action labels.
We select action labels for the entire dataset manually, using the VIA annotation tool~\cite{dutta2019vgg}. Fig.~\ref{fig:ground_truth} shows some annotation examples.

\vspace{-1.5mm}
\section{The H2O Dataset}\label{sec:dataset}

\label{sec:dataset_overview}
We acquired the images of the H2O dataset in indoor settings in which the subjects interact with eight different objects using both of their hands. %
The dataset includes 571,645 RGBD frames, and features four participants performing 36 distinct action classes in three different environments. With the methodology described in Sec.~\ref{sec:method}, we annotate accurate ground-truth data for left and right hand pose, 6D object pose, camera pose and action labels. In our dataset we further provide MANO~\cite{romero2017embodied} hand fits for both left and right hand, and high-quality object meshes. In addition, we also compute scene point clouds using the camera poses and the synchronized RGBD data. Altogether, the curated dataset allows for a comprehensive understanding of the egocentric scene.
\vspace{-1mm}

\subsection{Capture Setup}
Fig.~\ref{fig:data_stat}(c) demonstrates our data capture setup. We use five Azure Kinect cameras to acquire synchronized RGB and depth images. To ensure synchronization between multiple cameras, we use physical cables between them. This results in less than $100$ microseconds of lag between cameras~\cite{azurekinect2020}. As instructed in~\cite{azurekinect2020}, to avoid interference between multiple depth cameras, we further offset camera captures from one another by $160$ microseconds, which results in a total maximum of only 0.74 ms of delay between cameras. We place four different static cameras at arbitrary locations that cover hand-object interactions. An egocentric camera is further mounted on the forehead of a helmet and adjusted by the participants to set egocentric views. We calibrate all of the five cameras with nine IR reflective balls as explained in Sec.~\ref{sec:camera_pose_estimation}. The data is acquired in three  environments (\eg hall, office and kitchen) using several different backgrounds. We record videos at a resolution of 1280x720 pixels for both RGB and depth images with a frame rate of 30 fps. Each video corresponds to a series of actions involving various hand-object interactions. %

\subsection{Dataset Statistics}
\label{ssec:dataset_statistics}

We divide the dataset into a training and test set. We split the training and test data with a subject-based split where we leave one subject out for testing and the rest for training. We further use a part of the training data of one subject as the validation dataset for model selection. The data from multiple views consists of 344,645 frames for training, 73,380 frames for validation and 153,620 frames for testing.

\begin{figure}[t]
\begin{center}
\hspace{-8.3mm}{\includegraphics[width=1.08\columnwidth]{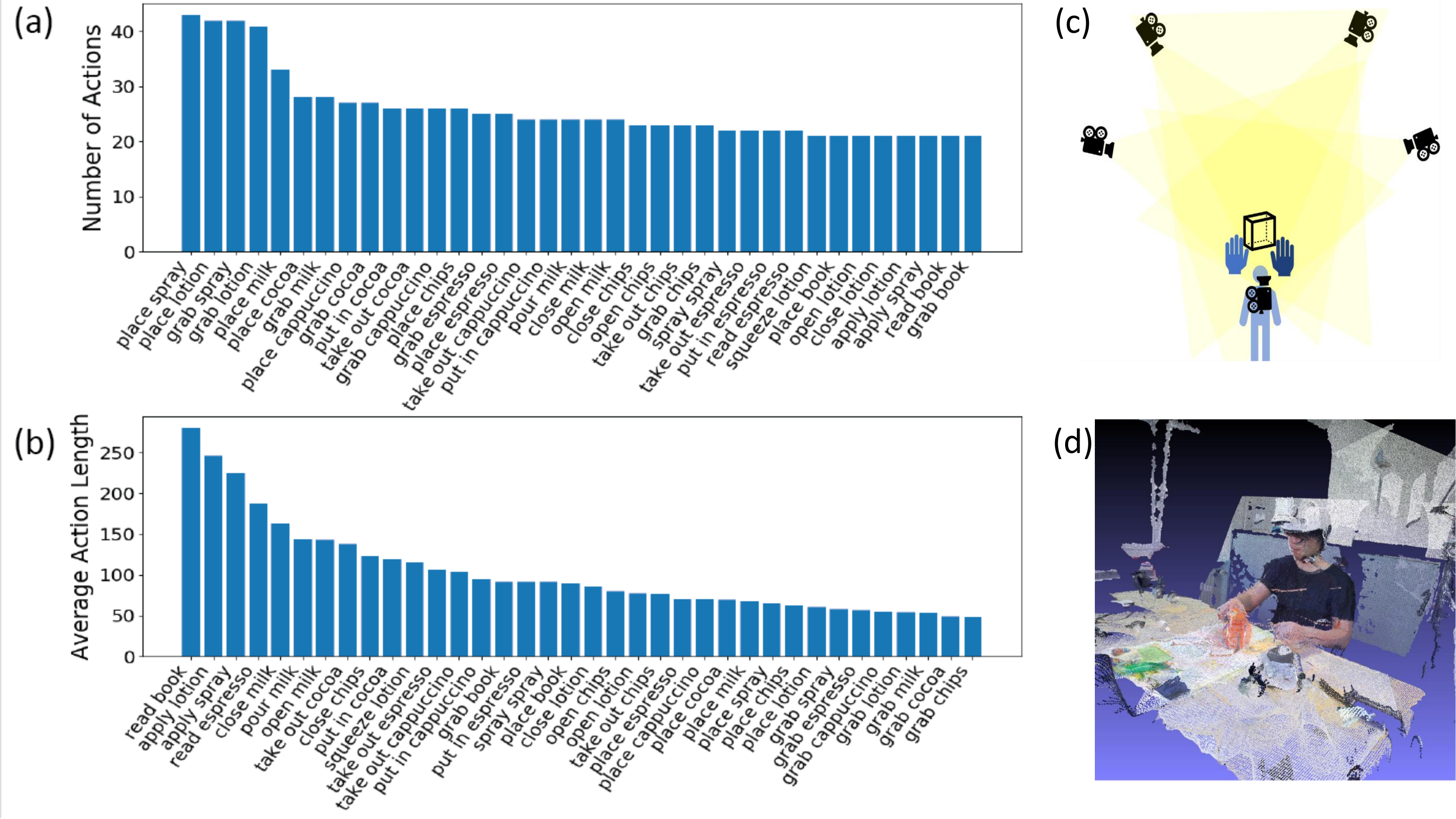}}
\end{center}
\vspace{-7mm}
\caption{\small (a) Number of instances per action in the H2O dataset. (b) Average number of frames per each action class. (c) Schematic camera capture setup. Four  static cameras can capture parts that can not be observed by the egocentric view. (d) Scene point cloud computed from multi-view data.}
\label{fig:data_stat}
\vspace{-6mm}
\end{figure}

We plot the number of instances per action and the average number of frames for each action class in Fig.~\ref{fig:data_stat}. Action instances are well distributed across the dataset with the least frequent action appearing 21 times. In the dataset, both hands are used in 57.8\%, only left hand in 12.4\%, and only right hand in 29.8\% of the dataset. The length of action clips spans a wide range demonstrating the diversity of the dataset that includes both slow and fast actions. 

\vspace{-2mm}
\section{Recognizing 3D Hand-Object Interactions}\label{sec:model}

Given the rich annotations of H2O, our goal is to construct comprehensive interpretations of egocentric scenes from image sequences to understand human interactions. For this purpose, we propose a unified framework that jointly estimates the poses of two hands \& the manipulated objects, and recognizes egocentric interactions. We use this framework to establish baselines on first person interaction recognition and hand \& object pose estimation.

\newcommand{\bx}[0]{\mathbf{x}}
\newcommand{\by}[0]{\mathbf{y}}
\boldparagraph{Pose Prediction.} We build upon the network architecture of~\cite{tekin2019h+} to estimate the poses of both left and right hand, and the pose of the manipulated object. While~\cite{tekin2019h+} addresses only single hand scenarios, in our case, we aim to predict the pose of both hands. To this end, each frame in a sequence is passed through a fully convolutional network with a backbone of YOLOv2~\cite{redmon17}. We produce a 3D grid as the output of our fully convolutional network, instead of producing a 2D grid as in~\cite{redmon17}. To be able to predict the pose of both hands and objects at the same time, we associate each output grid cell with 3 vectors for left hand, right hand and the manipulated object. These vectors contain target values for left hand (${\by}^{h,l}_{i}$), right hand (${\by}^{h,r}_{i}$) and object pose (${\by}^{o}_{i}$), with overall confidence values (${c}^{h,l}$,${c}^{h,r},{c}^{o}$) for individual pose predictions. The confidence values are defined on-the-fly during training as a function of the distance of the predicted poses to the ground-truth ones. The final layer of our single-shot network produces, for each cell $i$, predictions for left hand ($\hat{\by}^{h,l}_{i}$), right hand ($\hat{\by}^{h,r}_{i}$) and object ($\hat{\by}^o_{i}$),
along with their associated overall confidence values, $\hat{c}^{h,l}_{i}$, $\hat{c}^{h,r}_{i}$ and $\hat{c}^o_i$.
For each frame, the loss function to train our network is defined as follows:

\begin{footnotesize}
	\begin{align}
		\mathcal{L} & =   \lambda_{pose} \sum_{i} (|| \hat{\by}^{h,l}_{i} - \by^{h,l}_{i} || + || \hat{\by}^{h,r}_{i} - \by^{h,r}_{i} || + || \hat{\by}_{i}^o - \by^o_{i} ||) \\
		            & + \lambda_{conf} \sum_{i} ( (\hat{c}^{h,l}_i - c^{h,l}_i)^2 + (\hat{c}^{h,r}_i - c^{h,r}_i)^2 + (\hat{c}^o_i - c^o_i)^2 )
	\end{align}
\end{footnotesize}

\vspace{-2mm}

\;\; While the poses for the left and right hand are defined by 3D joint coordinates, object pose is parameterized by corner points of a 3D bounding box surrounding the object. Given the control point predictions of the network on the 3D bounding box, 6D object pose can be efficiently computed by aligning the predictions to the reference 3D bounding box with a rigid transformation. Predictions with low confidence values are pruned and the ones with high confidence values are selected as pose predictions.
\vspace{-2mm}

\begin{figure}[t]
	\centering
	\captionsetup[subfigure]{labelformat=empty}
	\vspace{-4mm}
	\subfloat[]{\includegraphics[width=0.45\columnwidth]{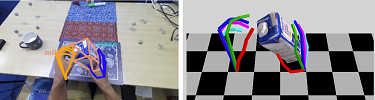}}\hspace{2mm}
	\subfloat[]{\includegraphics[width=0.45\columnwidth]{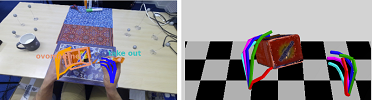}} \\
	\vspace{-6mm}
	\subfloat[]{\includegraphics[width=0.45\columnwidth]{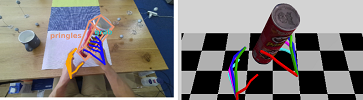}}\hspace{2mm}
	\subfloat[]{\includegraphics[width=0.45\columnwidth]{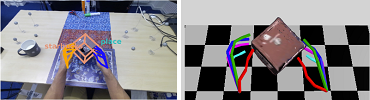}} \\
	\vspace{-6mm}
	\subfloat[]{\includegraphics[width=0.45\columnwidth]{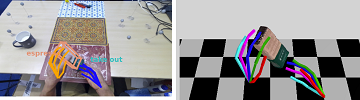}}\hspace{2mm}
	\subfloat[]{\includegraphics[width=0.45\columnwidth]{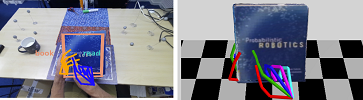}}
	\vspace{-9mm}
	\caption{\small Qualitative results on the $H2O$ dataset. We show estimated hand 3D pose, object 6D pose, and action labels. The proposed method can properly handle challenging occlusions.
	}
	\label{fig:results}
	\vspace{-6mm}
\end{figure}

\boldparagraph{Interaction Recognition.} RNNs have been successfully used before to recognize actions~\cite{baradel2018object,tekin2019h+}. However they do not fully leverage the special graph structure of the skeleton data for hand-object interactions. Therefore, we resort to parameterizing the left hand skeleton, right hand skeleton and object bounding box as individual graphs and combine them in a multi-graph structure. We then
compute the topology of the multi-graph structure using a graph convolutional network (GCN) by learning the links across hand and object locations that are involved in interaction. While modeling intra-dependencies within a single graph, this framework also allows for learning interdependencies between left hand-right hand, left hand-object, and right hand-object.

More particularly, we employ a spatiotemporal graph to encode both spatial and temporal information as in ST-GCN~\cite{yan2018spatial} and 2s-AGCN~\cite{shi2019two}. Standard ST-GCN~\cite{yan2018spatial} for human action recognition models structured information between body skeleton joints using

\vspace{-2mm}

\begin{small}
	\begin{equation}
		{\bf f_{out}} = \sum_j {\bf W_j} {\bf f_{in}} ({\bf A_j} \odot {\bf M_j})
		\label{eq:stgcn}
	\end{equation}
\end{small}

\vspace{-2mm}

\noindent where ${\bf f_{in}} \in \mathbb{R}^{C_{in} \times T \times N}$ is an input feature map, ${\bf A_j} \in \mathbb{R}^{N\times N}$ is an adjacency matrix that represents skeletal connections,  ${\bf W_j} \in \mathbb{R}^{C_{out} \times C_{in} \times 1 \times 1}$ is a weight vector of $1 \times 1$ convolutions and ${\bf M_j} \in \mathbb{R}^{N \times N}$ is an  attention map. Here $j$ denotes the vertex neighborhood defined by the convolutional kernel, $C$ is the number of channels, $T$ is the temporal length and $N$ is the number of vertices. ST-GCN works on a single graph entity, $\eg$ human skeleton, and models intra-skeleton connections with a fixed adjacency matrix. In our case, in addition to intra-graph dependencies, we aim to model also inter-graph dependencies between hands and objects. Since each time different hand and object parts are involved in interactions, a fixed adjacency matrix to model inter-dependencies would not yield optimal results. Therefore, individually for left hand, right hand and object,  we employ the following to be able to model their dependencies:

\vspace{-3mm}

\begin{footnotesize}
	\begin{equation}
		\hspace{-3mm}{\bf f_{out}} = \sum_j {\bf W_j} {\bf f_{in}} ({\bf A_{j,intra}} + {\bf A_{j,inter}} + {\bf T_{j,intra}} + {\bf T_{j,inter}} + {\bf S_j})
		\label{eq:topology-aware}
	\end{equation}
\end{footnotesize}

\vspace{-3mm}

\;\; While ${\bf A_{j, intra}}$ plays the same role as ${\bf A_j}$ in Eq.~\ref{eq:stgcn} for left hand, right hand and object, ${\bf A_{j, inter}}$ models inter-related dependencies between hands and objects via static connections between symmetric hand parts and object center. Here, both of these matrices are fixed adjacency matrices as in ST-GCN (Eq.~\ref{eq:stgcn}). In addition to them, we represent inter-connections between left hand and right hand, left hand and object, and right hand and object with an additional adjacency matrix, ${\bf T_{j,inter}}$. Differently from ${\bf A_{j,inter}}$, $\bf T_{j,inter}$ is not fixed, but rather parameterized. Its values are unconstrained and jointly optimized with other network parameters, which means that the graph topology and edge weights are fully learned from the training data. In addition to ${\bf T_{j,inter}}$, we also use an additional parameterized adjacency matrix, ${\bf T_{j,intra}}$ that adaptively learns intra-related dependencies within single graph entities (\eg left hand, right hand or object) during interaction. This data-driven model allows us to learn graphs that are fully targeting the hand-object interaction task.

Note that in contrast to~Eq.~\ref{eq:stgcn}, we do not use an attention map as in~\cite{shi2019two}, since our \emph{parameterized} adjacency matrices can play the same role of the attention mechanism performed by ${\bf M_j}$ in Eq.~\ref{eq:stgcn} to attribute more importance to edges between hands and objects that are involved in interaction. Besides in Eq.~\ref{eq:stgcn},  if one of the elements of ${\bf A_{j}}$ is $0$, the result  will be 0 regardless of the value of ${\bf M_j}$ due to the dot multiplication. Therefore we use addition instead of dot multiplication in Eq.~\ref{eq:topology-aware} to allow for forming new connections between our graphs. Similarly with~\cite{shi2019two}, we use an additional data-dependent term, ${\bf S_j}$ in our formulation which learns a unique graph for each sample that use the dot product to measure the similarity of the two vertices in an embedding space.

\vspace{-2mm}
\begin{small}
	\begin{equation}
		{\bf S_j} = \emph{softmax}({\bf f_{in}^T} {\bf W_{\theta j}^T } {\bf W_{\phi j} } {\bf f_{in}})
	\end{equation}
\end{small}
\vspace{-2mm}

\noindent where ${\bf W_{\theta}}$ and ${\bf W_{\phi}}$  are the parameters of the embedding functions $\theta$ and $\phi$, respectively. Here, embedding functions are chosen as $1\times1$ convolutional layers.

By stacking the layers defined by Eq.~\ref{eq:topology-aware}, with a total of 10 layers, we build our topology-aware graph convolutional network (TA-GCN) for 3D interaction recognition. It takes at each iteration the combination of $\hat{\by}^{h,l}_{i}$, $\hat{\by}^{h,r}_{i}$ and $\hat{\by}^{o}_{i}$ as its initial feature map to model hand object-interactions. We demonstrate learned graph connections for a hand-object interaction scenario in Fig.~\ref{fig:connection_inter} and analyze our design choices in Sec. ~\ref{sec:evaluation}. We provide further details for the architecture, hyperparameters and training of the pose prediction and interaction recognition models in the Supp.~Mat..

\vspace{-1mm}
\section{Evaluation}\label{sec:evaluation}
In this section, we first verify the accuracy of our ground-truth annotations. We then present baseline results on hand \& object pose estimation and egocentric action recognition on our dataset. For the latter, we also compare our baseline approach against the state-of-the-art in action recognition and demonstrate the clear benefits of our approach based on hand-object poses with respect to the existing methods.

\vspace{-2mm}
\subsection{Dataset Analysis}
\boldparagraph{Verfication.} We verify the accuracy of our hand-object pose annotations on a random split of our dataset. To that end, we annotate 500 images on 5 different camera views with the fingertips of the hand and the predefined keypoints of the manipulated objects. We then triangulate these 2D points to get manual 3D annotations for hands and objects. We compute the distance of our annotations to those of the manually created ones to measure the accuracy of our poses. We demonstrate the results of our verification in Table~\ref{table:verification}. For both hands and the object, the error is approximately within a range of 1 cm, which  demonstrates the high precision of our dataset. Our error margin is comparable with those of~\cite{hampali2020honnotate, zimmermann2019freihand} even though our dataset features more mutual occlusions due to two-hand manipulation.

\begin{figure}[t]
	\begin{center}
		\hspace{-3mm}{\includegraphics[width=0.85\columnwidth]{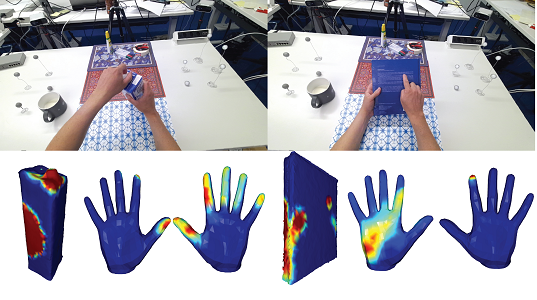}}
	\end{center}
	\vspace{-8mm}
	\caption{\small Contact modelling on H2O. Our dataset facilitates modelling hand-object contact and 3D affordances. }
	\vspace{-3mm}
	\label{fig:affordance}
\end{figure}

\begin{table}
	\begin{center}
		\begin{adjustbox}{width=0.9\columnwidth,center}
			\begin{tabular}{|c|c|c|c|}
				\hline
				Pose feature & Object           & Left hand        & Right hand       \\
				\hline
				Mean (std)   & 1.10 ($\pm$0.37) & 0.82 ($\pm$0.43) & 0.93 ($\pm$0.57) \\
				\hline
			\end{tabular}
		\end{adjustbox}
	\end{center}
	\vspace{-7mm}
	\caption{\small Hand \& object pose verification results (in cm) for evaluating the accuracy of the provided ground-truth data.}
	\vspace{-5mm}
	\label{table:verification}
\end{table}

\boldparagraph{Contact Modelling.} Having precise hand \& object pose annotations and meshes, H2O further facilitates modelling hand-object contact~\cite{brahmbhatt2020contactpose,taheri2020grab}. To this end, for each vertex in the hand mesh, we find the nearest vertices on the object within a certain a threshold (\eg $2$ cm). We then compute a histogram counting the number of neighbors for each vertex of the MANO mesh and normalize it to model contact hotspots on hand. We repeat the same procedure also for the object mesh to create a contact map on the object surface. We visualize example contact maps of our dataset in Fig.~\ref{fig:affordance}.

\subsection{Experimental Results}

\begin{figure}[t]
	\vspace{-3mm}
	\centering

	\subfloat[]{\includegraphics[width=0.33\columnwidth]{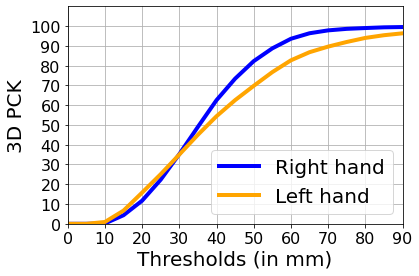}}
	\subfloat[]{\includegraphics[width=0.33\columnwidth]{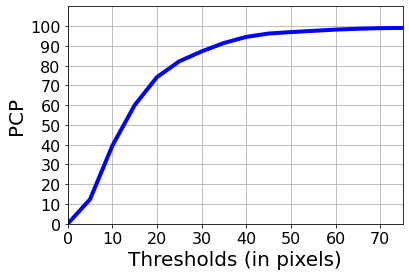}}
	\subfloat[]{\includegraphics[width=0.33\columnwidth]{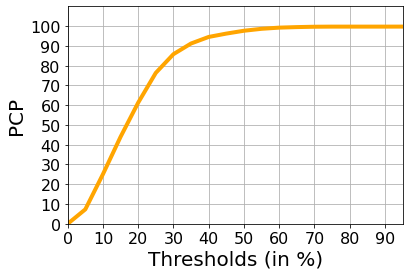}}
	\vspace{-4mm}
	\caption{\small Pose estimation results on the H2O dataset using different thresholds, for (a) hands with 3D PCK metric, and for objects with (b) 2D reprojection and (c) ADD metric.}\vspace{-1mm}
	\label{fig:hpecomparison}
	\vspace{-4mm}
\end{figure}

\boldparagraph{Predicting jointly the 3D pose of two hands and the manipulated objects.} We train and evaluate our method using the training, validation and test splits described in Sec.~\ref{ssec:dataset_statistics} and report baseline pose estimation accuracies for hands and objects in Fig.~\ref{fig:hpecomparison}. We use the percentage of correctly estimated poses to evaluate hand and object pose estimation accuracy. Specifically, we use the 3D PCK metric for hand pose estimation, and the 2D reprojection and ADD metrics for object pose estimation as in~\cite{tekin2019h+}.
We demonstrate that our method can reliably predict the pose of two hands and the manipulated objects with a low error margin and constitutes a strong baseline for joint pose estimation of two hands interacting with objects. Note also that our approach constitutes the first method and baseline for estimating the pose of \emph{two hands interacting with objects} from a single RGB image. We still evaluate our approach against single hand-object pose estimation methods of~\cite{hasson2020leveraging, tekin2019h+}, for comparison purposes, in Table~\ref{table:baselines} and further provide qualitative examples of our pose predictions in Fig.~\ref{fig:results}.

\begin{table}[t]

	\centering
	\tabcolsep=0.02cm
	\scalebox{0.62}{
		\begin{tabular}[b]{lclc}
			\hline
			Model                & Acc. (\%) \\
			\hline
			\textsc{Left Hand}   & 33.61     \\
			\textsc{Object}      & 48.55     \\
			\textsc{Right Hand } & 52.70     \\
			\textsc{Both Hands}  & 58.92     \\
			\textsc{All}         & 79.25     \\
			\hline
		\end{tabular}
		\quad
		\begin{tabular}[b]{lclc}
			\hline
			Model                         & Acc. (\%) \\
			\hline
			\textsc{No Interconnection}   & 75.52     \\
			\textsc{Left Hand-Right Hand} & 76.76     \\
			\textsc{Hands-Object}         & 78.84     \\
			\textsc{All Interconnections} & 79.25     \\
			\hline
		\end{tabular}
		\begin{tabular}[b]{lclc}
			\hline
			Model                                   & Acc. (\%) \\
			\hline
			\textsc{ST-GCN}                         & 73.86     \\
			\textsc{TA-GCN} wo ${\bf S_j}$          & 73.44     \\
			\textsc{TA-GCN} wo ${\bf T_{j, inter}}$ & 75.52     \\
			\textsc{TA-GCN} wo ${\bf T_{j, intra}}$ & 76.76     \\
			\textsc{TA-GCN} wo ${\bf A_{j, inter}}$ & 76.35     \\
			\textsc{TA-GCN} wo ${\bf A_{j, intra}}$ & 77.59     \\

			\textsc{TA-GCN}                         & 79.25     \\
			\hline
		\end{tabular}
	}
	{ \small \quad\quad\quad\quad (a) \hspace{2.6cm} (b) \hspace{2.6cm} (c) }
	\vspace{-3mm}
	\caption{\small Impact of different (a) input modalities, (b) interconnections and (c) graph terms on interaction recognition accuracy.}
	\label{tab:action_ablation_study}
	\vspace{-3mm}
\end{table}

\begin{figure}[t]
	\vspace{-6mm}
	\begin{center}
		{\includegraphics[width=0.95\columnwidth]{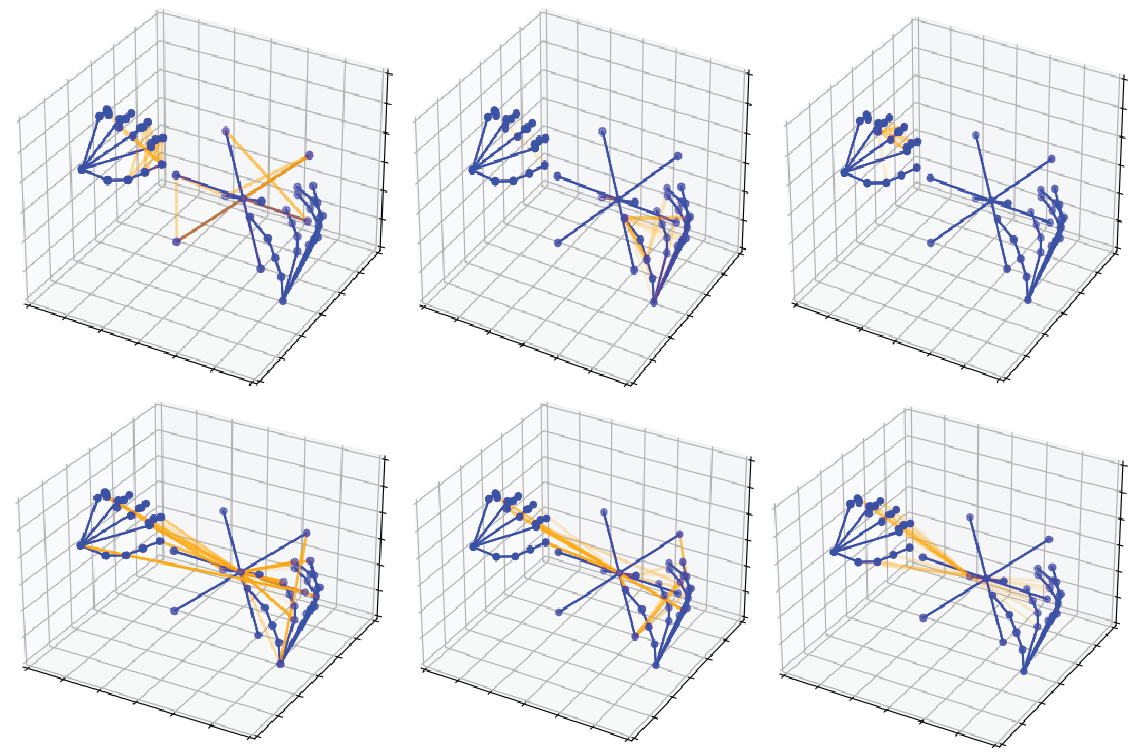}}
	\end{center}
	\vspace{-7mm}
	\caption{\small Learned graph connections for different layers. We demonstrate the top-20 learned  intra-(top) and  inter-(bottom) connections at layer 1, 5 and 9, respectively, in each column. The thickness of the connections corresponds to the weight of learned connection value. Hand-object connections are given more weight than hand-hand connections during interaction with an object. Our model attributes more importance to fingertips and DIP joints that are more commonly involved in manipulation. }
	\label{fig:connection_inter}
	\vspace{-2mm}
\end{figure}

\boldparagraph{Interaction recognition.} In Table~\ref{tab:action_ablation_study}(a), we show the influence of different input modalities on the accuracy of interaction recognition on the H2O dataset. To this end, we evaluate the impact of hand \& object poses for interaction recognition. Hand pose and object keypoints are predicted through our single pass network described in Sec.~\ref{sec:model}.
We show that the combination  of right and left hand pose as well as the combination of hand and object poses significantly improve overall action recognition scores, which demonstrates the individual contributions and the complementary nature of each input modality. We further evaluate the importance of modelling inter-dependencies between both hands and objects in Table~\ref{tab:action_ablation_study}(b) and demonstrate that modelling interdependencies between left hand \& right hand, and hands \& objects, boosts the accuracy for recognizing interactions.  In Table~\ref{tab:action_ablation_study}(c), we evaluate the influence of different terms of Eq.~\ref{eq:topology-aware} and demonstrate that with all the graphs added together, the model obtains the best results compared to the baselines. We visualize the learned connections of our model in Fig.~\ref{fig:connection_inter}.

We further compare our action recognition accuracy to the state-of-the-art image-based learning methods of C2D~\cite{wang2018non},  I3D~\cite{carreira2017quo} and SlowFast~\cite{feichtenhofer2019slowfast} using the PySlowFast library~\cite{fan2020pyslowfast} and pose-based learning methods of H+O~\cite{tekin2019h+} and ST-GCN~\cite{yan2018spatial} and show our results in Table~\ref{table:baselines}. Following~\cite{fan2020pyslowfast}, we train image-based models using a batch size of 16 and a temporal window size of 64 frames with a sampling ratio of 2. We use a ResNet-50 backbone and train the network using SGD with a learning rate of $0.1$. Pose-based methods are trained as in~\cite{tekin2019h+,yan2018spatial}, and evaluated with the estimated poses using our method from RGB images of our dataset. Our approach to interaction recognition achieves the highest validation and test accuracy on the H2O dataset, demonstrating the effectiveness of our method and the importance of the 3D pose predictions facilitated by H2O.

\begin{table}
	\vspace{-1mm}
	\begin{center}
		\scalebox{0.62}{
			\begin{tabular}{lccc}
				\hline
				\textbf{Method}                   & Left h.        & Right h.       & Object         \\

				\hline
				Hasson\cite{hasson2020leveraging} & \textbf{39.56} & -              & 67.47          \\ %
				Hasson\cite{hasson2020leveraging} & -              & 41.87          & 66.05          \\ %
				H+O\cite{tekin2019h+}             & 41.42          & -              & 48.06          \\ %
				H+O\cite{tekin2019h+}             & -              & 38.86          & 52.57          \\ %
				Ours                              & 41.45          & \textbf{37.21} & \textbf{47.90} \\
				\hline
			\end{tabular}
			\begin{tabular}{l  c c}
				\hline
				Model                                     & Val acc. (\%)  & Test acc. (\%) \\
				\hline\hline
				C2D \cite{wang2018non}                    & 76.10          & 70.66          \\
				I3D \cite{carreira2017quo}                & 85.15          & 75.21          \\
				SlowFast \cite{feichtenhofer2019slowfast} & 86.00          & 77.69          \\
				\hline
				H+O \cite{tekin2019h+}                    & 80.49          & 68.88          \\
				ST-GCN \cite{yan2018spatial}              & 83.47          & 73.86          \\
				OURS (TA-GCN)                             & \textbf{86.78} & \textbf{79.25} \\

				\hline
			\end{tabular}
		}
	\end{center}
	\vspace{-7mm}
	\caption{\small  Pose errors (left, in mm) and action accuracies (right). Single hand methods of~\cite{hasson2020leveraging,tekin2019h+} are separately trained for left \& right hand. \cite{tekin2019h+,yan2018spatial} use pose predictions of our method.}
	\label{table:baselines}
	\vspace{-4mm}
\end{table}

\section{Conclusion}\label{sec:conclusion}

In this paper, we propose a method to collect a dataset of two hands manipulating objects for first person interaction recognition. We provide a rich set of annotations including action labels, object classes, 3D left \& right hand poses, 6D object poses, camera poses and scene point clouds. We further propose the first method to jointly recognize the 3D poses of two hands manipulating objects and a novel topology-aware graph convolutional network for recognizing hand-object interactions. Our framework models the interactions between hands and objects in 3D to recognize actions from first-person views and yields state-of-the-art accuracy.  We believe that our dataset and experiments can be of interest to communities of 3D hand pose estimation, 6D object pose estimation, hand-object interaction, robotics and action recognition, and help bridge the gap between hand-object interaction and egocentric action recognition.

\noindent \textbf{Acknowledgements.} Taein Kwon was supported by the Microsoft Mixed Reality \& AI Z\"urich Lab PhD scholarship. 
The authors thank Silvano Galliani, Joshua Elsdon, Yana Hasson, Jeff Delmerico, Helen Oleynikova and Mihai Dusmanu for helpful discussions.

{\small
	\bibliographystyle{ieee_fullname}
	\bibliography{main}
}

\clearpage

\setcounter{table}{0}
\renewcommand{\thetable}{S\arabic{table}}%
\setcounter{figure}{0}
\renewcommand{\thefigure}{S\arabic{figure}}%
\setcounter{section}{0}
\renewcommand{\thesection}{S}%

\title{Supplementary Material: \\H2O: Two Hands Manipulating Objects for First Person Interaction Recognition}

\author{Taein Kwon$^1$, Bugra Tekin$^2$, Jan St{\"u}hmer$^{3*}$, Federica Bogo$^1$, and Marc Pollefeys$^{1,2}$\\
	$^1$ETH Zurich, $^2$Microsoft, $^3$Samsung AI Center, Cambridge
}

\maketitle
\thispagestyle{empty}

In the supplemental material, we provide further analysis of our annotation method and evaluate different error and regularization terms. Next, we explain how the training images were prepared for object pose estimation. We then provide the implementation details, evaluation metrics and further analysis of our method for joint pose estimation and interaction recognition. We  finally  present  further  qualitative  results of our method.

\subsection{Analysis of the Annotation Method}

\boldparagraph{Influence of different error terms.} In Table~\ref{table:ablation_veri}, we analyze the influence of different error terms in our joint loss function for annotating hand \& object poses. To validate the accuracy of our pose estimates, we annotate the fingertips of the hands on 500 images from 5 different views. We start with the silhouette error term, $\mathcal{L}_{s}$, since it optimizes the shape of the hands. We then progressively add to our loss function, the 2D joint error term ($\mathcal{L}_{2D}$), the 3D joint error term ($\mathcal{L}_{3D}$), the physical constraint error term ($\mathcal{L}_{phy}$), and the 3D mesh surface error term ($\mathcal{L}_{m}$). We observe that $\mathcal{L}_{2D}$ and $\mathcal{L}_{3D}$ significantly increase joint estimation accuracy. While $\mathcal{L}_{m}$ improves the estimates for subtle hand mesh shape and location, the improvement in joint accuracy is less pronounced. $\mathcal{L}_{phy}$ improves both the physical plausibility and the accuracy of pose annotations. Further smoothing and pose corrections give an additional boost in accuracy. Overall, all the terms of our optimization function in Eq.~\ref{eq:supp_loss_sum} increase the quality of our pose estimates.

\vspace{-4mm}
\begin{equation}\label{eq:supp_loss_sum}
	\begin{aligned}
		\hat \theta_{f} = \argmin_\theta\sum_{c=1}^{N_{C}}(\lambda_{1}\mathcal{L}_{s}+\lambda_{2}\mathcal{L}_{2D})+\lambda_{3}\mathcal{L}_{3D}+ \\ \lambda_{4}\mathcal{L}_{p}+\lambda_{5}\mathcal{L}_{phy}+\lambda_{6}\mathcal{L}_{a}+\lambda_{7}\mathcal{L}_{m}
	\end{aligned}
	\vspace{-2mm}
\end{equation}

We provide below additional details for the terms of our loss function.

\boldparagraph{Silhouette error term.} We use object masks obtained using a self-trained Mask RCNN~\cite{he2017maskrcnn}. %
For hands, we estimate hand joint 2D locations in RGB using OpenPose~\cite{cao2018openpose}, and use them to initialize the GrabCut algorithm~\cite{rother2004grabcut}. For each camera $c$, we merge the hand mask obtained via GrabCut with the object mask into a single mask, $M_{c,h,o}$, and define our silhouette error term as:
\vspace{-2mm}
\begin{equation}\label{eq:loss_mask}
	\begin{aligned}
		\mathcal{L}_{s}(\theta)=\sum_{i=1}^{N_{V}}\norm{M_{c,h,o}[j]-\Pi_c(H_V(\theta)[i])} \\
		\text{where } j = \argmin_{j}\norm{M_{c,h,o}[j]-\Pi_c(H_V(\theta)[i])}
	\end{aligned}
	\vspace{-2mm}
\end{equation}
where $||\cdot||$ denotes the 2-norm, $\Pi_{c}(\cdot)$ gives the 2D projection of a 3D point onto the image plane, $M_{c,h,o}[j]$ returns the $j$th coordinate in the mask of the $c$th camera, and $H_V(\theta)[i]$ returns the $i$th vertex of the hand mesh. We compute Eq.~\ref{eq:loss_mask} for each camera.

\boldparagraph{Physical constraint regularization.} To avoid physically invalid poses (e.g. a finger inside an object), we regularize our loss function with an additional term as in~\cite{hasson2019learning}:

\begin{equation}\label{eq:loss_phy}
	\begin{aligned}
		\mathcal{L}_{phy}(\theta_{h},\theta_{o})=\lambda_{r}\mathcal{L}_{R}+(1-\lambda_{r})\mathcal{L}_{A}
	\end{aligned}
\end{equation}
where $\theta_{h}$ are hand pose, $\theta_{o}$ are object pose parameters, $\mathcal{L}_a$ is attraction loss and, $\mathcal{L}_r$ is repulsion loss. While repulsion loss penalizes interpenetration of hand and objects, attraction loss penalizes the cases in which hand vertices are in the vicinity of the objects but the surfaces are not in contact. In our experiments, we set $\lambda_{r}$ to $0.8$.

\begin{table*}
	\begin{center}
		\begin{adjustbox}{width=\linewidth,center}
			\begin{tabular}{|c|c|c|c|c|c|c|c|}
				\hline
				Terms            & $\mathcal{L}_{s}$ & $\mathcal{L}_{s}+\mathcal{L}_{2D}$ & $\mathcal{L}_{s}+\mathcal{L}_{2D}+\mathcal{L}_{3D}$ & \makecell{$\mathcal{L}_{s}+\mathcal{L}_{2D}+$                                                          \\$\mathcal{L}_{3D} +\mathcal{L}_{m}$} & \makecell{$\mathcal{L}_{s} + \mathcal{L}_{2D}+$ \\ $\mathcal{L}_{3D}+\mathcal{L}_{phy}$} & \makecell{$\mathcal{L}_{s} +\mathcal{L}_{2D}+\mathcal{L}_{3D}+$ \\ $\mathcal{L}_{phy}+\mathcal{L}_{m}$} & \makecell{$\mathcal{L}_{s}+\mathcal{L}_{2D}+\mathcal{L}_{3D}+$ \\ $\mathcal{L}_{phy}+\mathcal{L}_{m} + Smoothing $}\\
				\hline
				Left Mean (std)  & 2.87 ($\pm$1.48)  & 1.25 ($\pm$1.12)                   & 1.09 ($\pm$1.03)                                    & 1.08 ($\pm$1.02)                              & 0.95 ($\pm$0.79) & 0.95 ($\pm$0.77) & 0.82 ($\pm$0.43) \\
				Right Mean (std) & 3.02 ($\pm$1.65)  & 1.31 ($\pm$1.03)                   & 1.11 ($\pm$0.98)                                    & 1.11 ($\pm$1.00)                              & 1.05 ($\pm$0.98) & 1.04 ($\pm$0.94) & 0.93 ($\pm$0.57) \\
				\hline
			\end{tabular}
		\end{adjustbox}
	\end{center}
	\vspace{-5mm}
	\caption{\small Impact of different error terms in our joint loss function. Errors are given in millimeters (mm). Note that all the experiments include regularization terms, $\mathcal{L}_{a}$ and $\mathcal{L}_{p}$, to avoid unrealistic hand poses.}
	\label{table:ablation_veri}
	\vspace{-4mm}
\end{table*}

\boldparagraph{Hand joint angle limit regularization.} In our loss function, we further penalize unrealistic joint angles as in~\cite{hampali2020honnotate}:
\begin{equation}\label{eq:loss_angle}
	\begin{aligned}
		\mathcal{L}_{a}(\theta)=\sum_{k=1}^{45} ( max(0,\underline{\theta_{a}}[k]-\theta_{a}[k])+ \\
		max(\theta_{a}[k]-\overline{\theta_{a}}[k],0))
	\end{aligned}
\end{equation}
where $\theta_{a}[k]$ is the $k_{th}$ joint angle, $\underline{\theta_{a}}$ is the lower limit of the angle, and $\overline{\theta_{a}}$ is the upper limit of the angle. There exist in total 45 joint angles. As also observed in~\cite{hampali2020honnotate}, the PCA space of the MANO hand model does not provide sufficient details to represent all possible hand poses. Therefore instead we use joint angle space which is more descriptive for hand poses. We calculate the limits of joint angles heuristically and give them in Table~\ref{table:hand_joint_angles}.

\boldparagraph{Pose prior.} In order to regularize hand pose, we model the distribution of hand poses provided in the MANO dataset~\cite{romero2017embodied} as a multivariate Gaussian. Based on this, we define a pose prior, $\mathcal{L}_{p}$, which penalizes the Mahalanobis distance between both left and right hand pose $\theta$ and the learned Gaussian distributions as in~\cite{romero2017embodied}. %

\vspace{-3mm}

\begin{equation}\label{eq:loss_prior}
	\begin{aligned}
		\mathcal{L}_{p}(\theta)=\sqrt{(\theta-\overline{\theta})^\top S^{-1}(\theta-\overline{\theta})}
	\end{aligned}
\end{equation}

\vspace{-1mm}

\noindent where S is the covariance of the hand pose distribution.

\begin{figure}
	\begin{center}
		\subfloat[]{\includegraphics[width=0.48\linewidth]{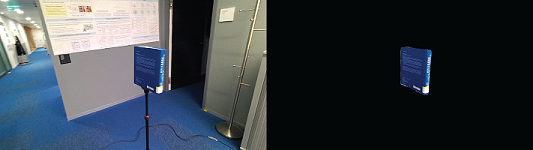}} \hspace{2mm} \subfloat[]{\includegraphics[width=0.48\linewidth]{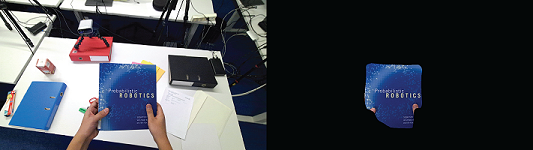}} \\
		\subfloat[]{\includegraphics[width=0.48\linewidth]{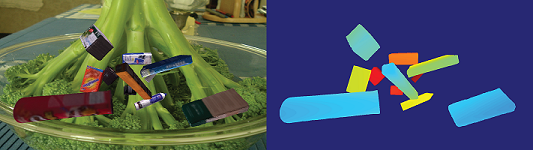}} \hspace{2mm}
		\subfloat[]{\includegraphics[width=0.48\linewidth]{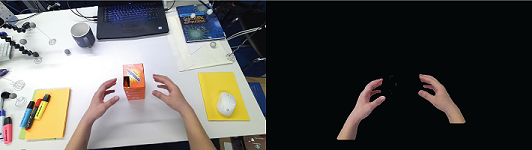}}
	\end{center}
	\vspace{-6mm}
	\caption{\small We show some examples of (a) our object images and their corresponding masks obtained by projecting the object models, (b) egocentric images and Mask R-CNN results, (c) synthetic images and the corresponding depth images for training DenseFusion, (d) hand masks obtained by GrabCut~\cite{rother2004grabcut}.}
	\label{fig:captured_data}
	\vspace{-2mm}
\end{figure}

\begin{table}[t]
	\begin{center}
		\begin{adjustbox}{width=\columnwidth,center}

			\begin{tabular}{l c c c c c c c c c c}
				\hline
				\multirow{2}{*}{Joint}    & \multicolumn{2}{c}{Index} & \multicolumn{2}{c}{Moddle} & \multicolumn{2}{c}{Pinky} & \multicolumn{2}{c}{Ring} & \multicolumn{2}{c}{Thumb}                                     \\\cline{2-3}\cline{4-5}\cline{6-7}\cline{8-9}\cline{10-11}
				                          & Min                       & Max                        & Min                       & Max                      & Min                       & Max  & Min  & Max  & Min   & Max  \\

				\hline
				\multirow{3}{*}{MPC(CMC)} & -0.45                     & 0                          & 0                         & 0                        & -1.5                      & 0.5  & -0.5 & 0.5  & 0     & 2    \\
				                          & -0.2                      & 0.2                        & -0.2                      & 0.2                      & -0.6                      & 0.6  & -0.4 & 0.4  & -0.66 & 0.83 \\
				                          & -0                        & 2                          & 0                         & 2                        & 0                         & 2    & 0    & 2    & 0     & 0.5  \\
				\hline
				\multirow{3}{*}{PIP(MCP)} & -0.3                      & 0.3                        & -0.3                      & 0.3                      & -0.3                      & 0.3  & -0.3 & 0.3  & -0.3  & 0.3  \\
				                          & 0                         & 0                          & 0                         & 0                        & 0                         & 0    & 0    & 0    & -1    & 1    \\
				                          & 0                         & 2                          & 0                         & 2                        & 0                         & 2    & 0    & 2    & 0     & 1    \\
				\hline
				\multirow{3}{*}{DIP(IP)}  & 0                         & 0                          & 0                         & 0                        & 0                         & 0    & 0    & 0    & 0     & 0    \\
				                          & 0                         & 0                          & 0                         & 0                        & 0                         & 0    & 0    & 0    & 0     & 0    \\
				                          & 0                         & 1.25                       & 0                         & 1.25                     & 0                         & 1.25 & 0    & 1.25 & 0     & 1    \\

				\hline
			\end{tabular}
		\end{adjustbox}
	\end{center}
	\vspace{-5mm}
	\caption{\small Hand joint limits we use in computing $\mathcal{L}_{a}$.}
	\vspace{-5mm}
	\label{table:hand_joint_angles}
\end{table}

\boldparagraph{Pose correction.} To provide an even higher quality for our pose annotations, we inspected our dataset after optimization and selected keyframes on our videos to generate smooth trajectories for hand \& object poses. The number of keyframes we choose is reported in Table~\ref{tab:key_frame_ratio}. We fix small errors of hand and object poses via interpolation based on these keyframes.

\boldparagraph{Implementation details of the annotation method.} We further provide implementation details for our annotation method below.

\begin{figure}[h]
	\vspace{-3mm}
	\begin{center}
		\subfloat[]{\includegraphics[width=0.7\linewidth]{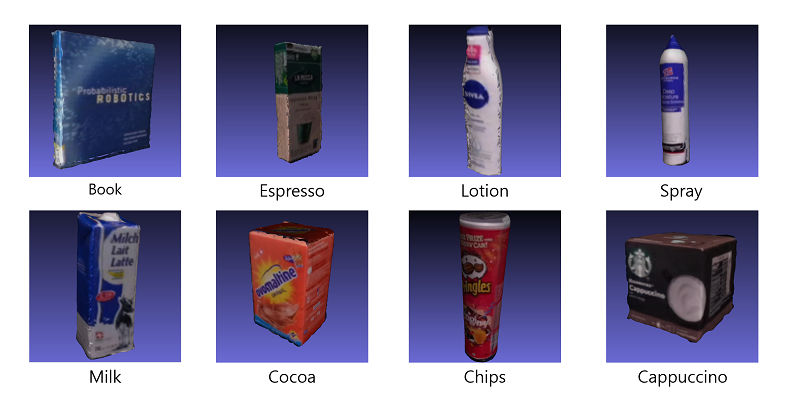}}
		\subfloat[]{\includegraphics[width=0.3\linewidth]{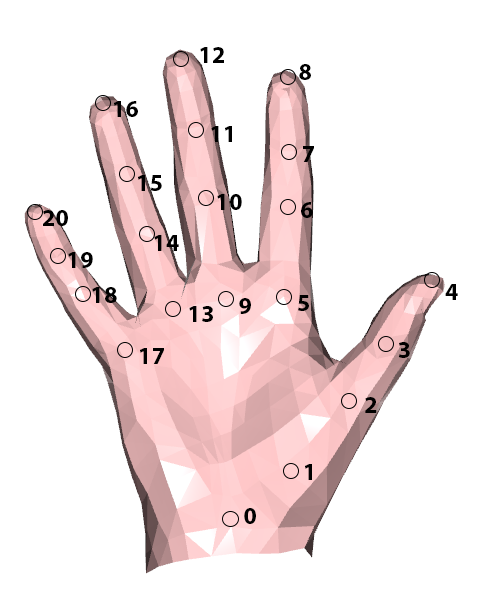}}
	\end{center}
	\vspace{-6mm}
	\caption{\small (a) Object 3D models obtained with our method. We reconstruct textured 3D meshes for $8$ different objects. (b) The map of the OpenPose~\cite{cao2018openpose} hand skeleton.}
	\label{fig:obj_meshes}
	\vspace{-5mm}
\end{figure}

\noindent \emph{Hand joint definition.} The MANO~\cite{romero2017embodied} model provides hand joints for 15 locations as shown in Table~\ref{table:hand_joint_angles}. In order to map hand joints from the  MANO skeleton to OpenPose~\cite{cao2018openpose} skeleton, we reorganize the order of joints and add wrist \& fingertip locations by selecting corresponding points on the MANO mesh as shown in Fig.~\ref{fig:obj_meshes}(b).

\begin{table}[h]
	\begin{center}
		\begin{adjustbox}{width=\linewidth,center}
			\scalebox{0.82}{
				\begin{tabular}{|c|c|c|c|c|}
					\hline
					                        & Object  & Left hand & Right hand \\
					\hline
					\# keyframes            & 107,300 & 88,342    & 88,264     \\
					\# interpolated  frames & 7,029   & 25,987    & 26,065     \\
					\hline
				\end{tabular}
			}
		\end{adjustbox}
	\end{center}
	\vspace{-5mm}
	\caption{\small The number of keyframes and interpolated frames for the annotations of hands and objects in our dataset.}
	\label{tab:key_frame_ratio}
	\vspace{-3mm}
\end{table}

\noindent\emph{Object pose estimation network.} We use DenseFusion~\cite{wang2019densefusion} on multi-view RGBD images to bootstrap our object pose annotations. To this end, we train on the RGB and depth images as well as the corresponding segmentation masks, given in Fig.~\ref{fig:captured_data}(a). We train the network using ADAM~\cite{kingma2014method} with a learning rate of 0.0001. We generate synthetic training images by superimposing object meshes (Fig.~\ref{fig:obj_meshes}(a)) with known 6D poses on random backgrounds and cover a large variety of object poses as shown in Fig.~\ref{fig:captured_data}(c). We create the object meshes using BADSLAM~\cite{schops2019bad}.

\noindent\emph{Segmentation masks.} To minimize silhouette error term and train DenseFusion, we require segmentation mask for objects. To generate segmentation masks, we train Mask R-CNN~\cite{he2017maskrcnn} with a ResNet-101 backbone. We optimize the network using SGD with a learning rate of 0.001. We obtain training data for masks by projecting our 3D models onto the images as shown in Fig.~\ref{fig:captured_data}~(a). An example result of Mask R-CNN is shown in Fig.~\ref{fig:captured_data}~(b). We use randomly selected COCO~\cite{lin2014microsoft} images for background augmentation (Fig.~\ref{fig:captured_data}~(c)). As discussed in the main paper, we use GrabCut to generate segmentation masks for hands. We provide an example segmentation mask for hands in Fig.~\ref{fig:captured_data}~(d).

\begin{figure*}[]
	\begin{center}
		\captionsetup[subfigure]{labelformat=empty}
		\subfloat[]{\includegraphics[width=1\textwidth]{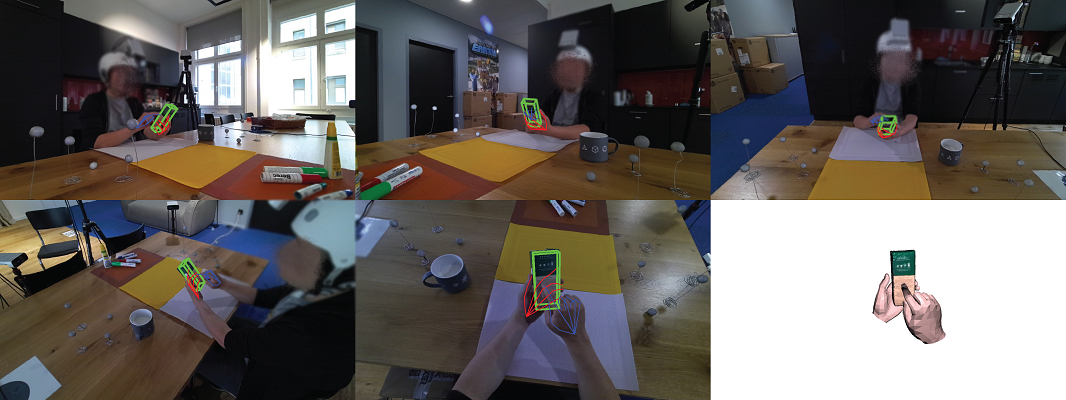}} \\
		\subfloat[]{\includegraphics[width=1\textwidth]{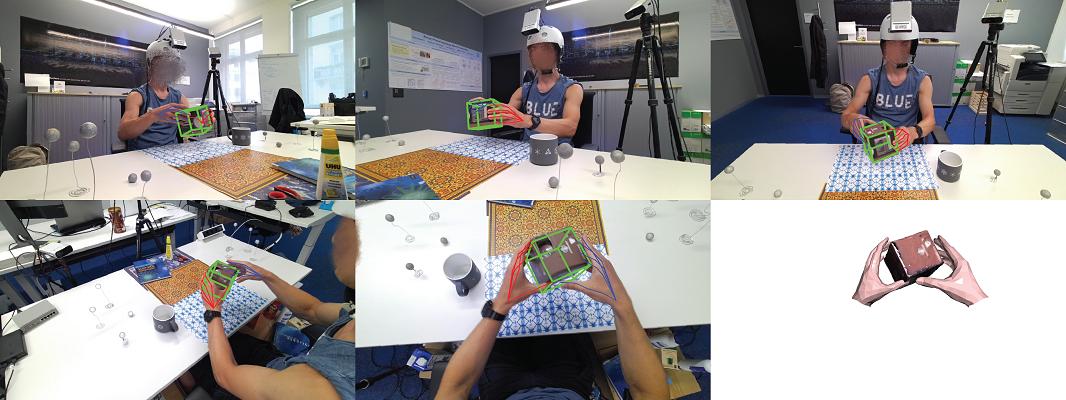}} \\
		\subfloat[]{\includegraphics[width=1\textwidth]{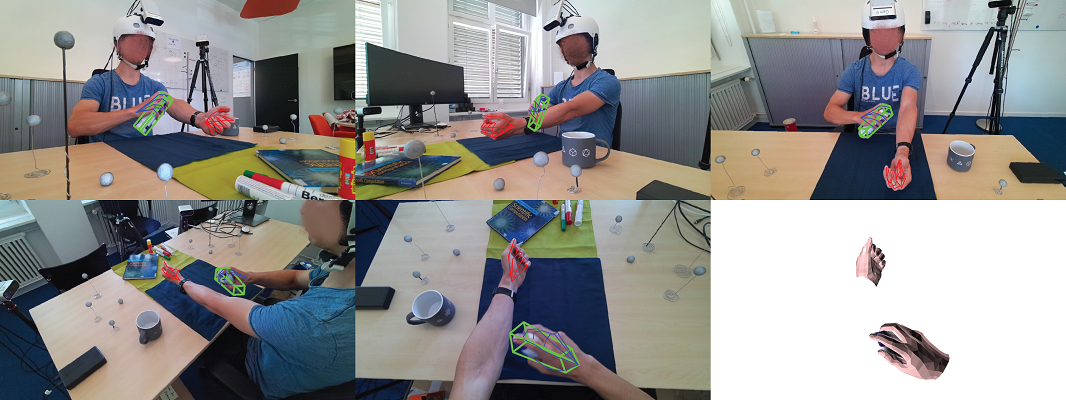}}

	\end{center}
	\caption{\small Some examples of ground-truth data of our dataset for hand \& object poses on five different camera views.}
	\label{fig:ground_truth1}
\end{figure*}

\begin{figure*}[]
	\begin{center}
		\captionsetup[subfigure]{labelformat=empty}
		\subfloat[]{\includegraphics[width=1\textwidth]{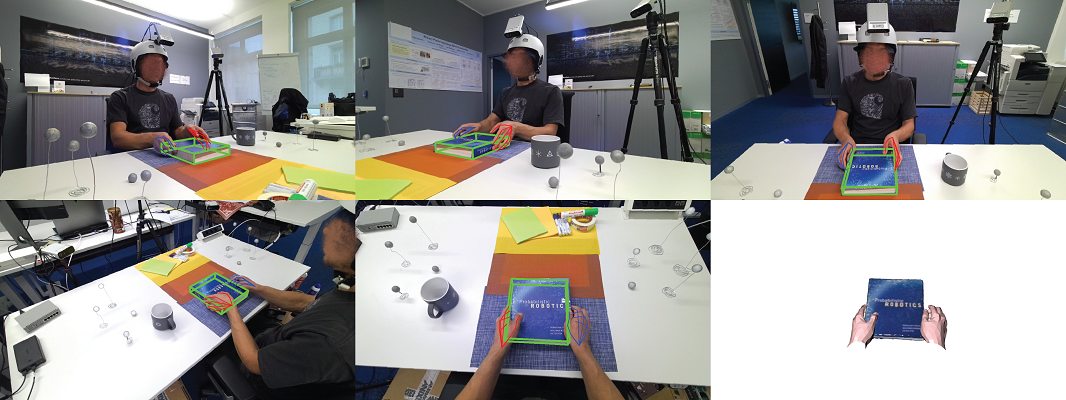}} \\
		\subfloat[]{\includegraphics[width=1\textwidth]{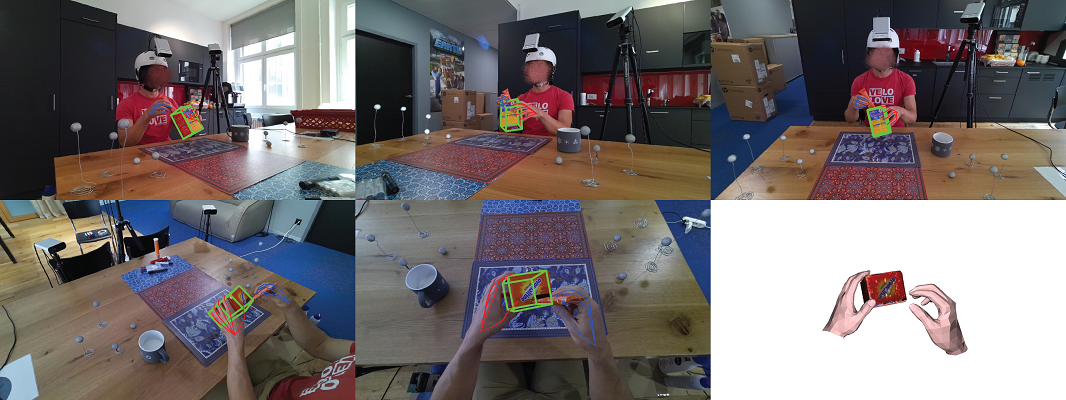}} \\
		\subfloat[]{\includegraphics[width=1\textwidth]{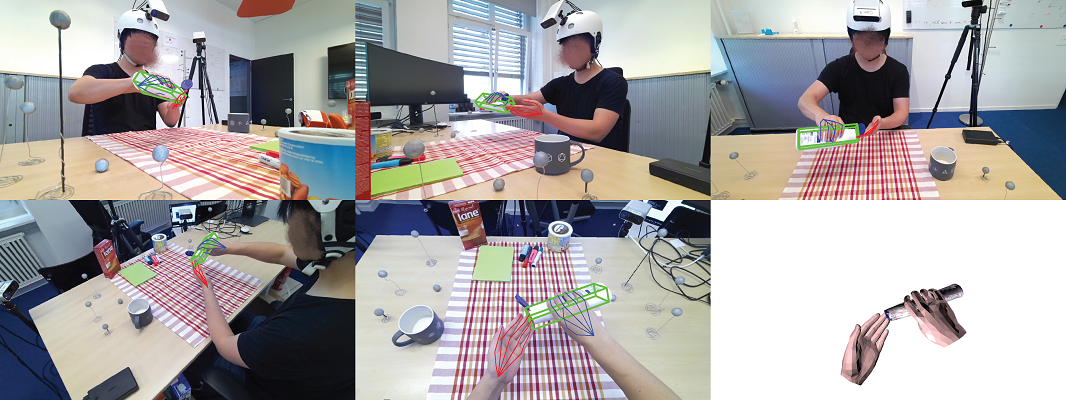}}

	\end{center}
	\caption{\small Some examples of ground-truth data of our dataset for hand \& object poses on five different camera views.}
	\label{fig:ground_truth2}
\end{figure*}

\boldparagraph{Example data.} We provide in Fig.~\ref{fig:ground_truth1} and Fig.~\ref{fig:ground_truth2} additional qualitative examples for our ground-truth data, which demonstrate the high fidelity and accuracy of our dataset.

\subsection{Analysis of Pose and Interaction Recognition}

\paragraph{Implementation details for pose prediction.} We provide in Table~\ref{tab:networkarchitecture} the full details of our network architecture. We use YOLOv2~\cite{redmon17} as the backbone of our network. The input to our network model is a $416\times416$ image. At the output layer, we produce a 3D grid instead of a 2D grid with a dimension of $13\times13\times5$, in width, height and depth axes, respectively. We set the grid cell size in image dimensions to $32\times32$ pixels and in depth dimension to $15$cm. We define the confidence of a prediction with a function that is inversely proportional to the distance of the prediction to the ground truth as in~\cite{tekin2019h+} with its default parameters. We use ADAM~\cite{kingma2014method} for optimization  with a learning rate of $0.0001$. We randomly change the hue, saturation and exposure of our images to augment our training data.

\begin{table}[t]
	\begin{center}
		\scalebox{0.55}{
			\begin{tabular}{|c|c|c|c|c|c|}
				\hline
				Layer & Type       & Filters & Size/Stride      & Input                          & Output                                                                       \\
				\hline
				0     & conv       & 32      & 3 $\times$ 3 / 1 & 416 $\times$ 416 $\times$    3 & 416 $\times$ 416 $\times$   32                                               \\
				1     & max        &         & 2 $\times$ 2 / 2 & 416 $\times$ 416 $\times$   32 & 208 $\times$ 208 $\times$   32                                               \\
				2     & conv       & 64      & 3 $\times$ 3 / 1 & 208 $\times$ 208 $\times$   32 & 208 $\times$ 208 $\times$   64                                               \\
				3     & max        &         & 2 $\times$ 2 / 2 & 208 $\times$ 208 $\times$   64 & 104 $\times$ 104 $\times$   64                                               \\
				4     & conv       & 128     & 3 $\times$ 3 / 1 & 104 $\times$ 104 $\times$   64 & 104 $\times$ 104 $\times$  128                                               \\
				5     & conv       & 64      & 1 $\times$ 1 / 1 & 104 $\times$ 104 $\times$  128 & 104 $\times$ 104 $\times$   64                                               \\
				6     & conv       & 128     & 3 $\times$ 3 / 1 & 104 $\times$ 104 $\times$   64 & 104 $\times$ 104 $\times$  128                                               \\
				7     & max        &         & 2 $\times$ 2 / 2 & 104 $\times$ 104 $\times$  128 & 52 $\times$  52 $\times$  128                                                \\
				8     & conv       & 256     & 3 $\times$ 3 / 1 & 52 $\times$  52 $\times$  128  & 52 $\times$  52 $\times$  256                                                \\
				9     & conv       & 128     & 1 $\times$ 1 / 1 & 52 $\times$  52 $\times$  256  & 52 $\times$  52 $\times$  128                                                \\
				10    & conv       & 256     & 3 $\times$ 3 / 1 & 52 $\times$  52 $\times$  128  & 52 $\times$  52 $\times$  256                                                \\
				11    & max        &         & 2 $\times$ 2 / 2 & 52 $\times$  52 $\times$  256  & 26 $\times$  26 $\times$  256                                                \\
				12    & conv       & 512     & 3 $\times$ 3 / 1 & 26 $\times$  26 $\times$  256  & 26 $\times$  26 $\times$  512                                                \\
				13    & conv       & 256     & 1 $\times$ 1 / 1 & 26 $\times$  26 $\times$  512  & 26 $\times$  26 $\times$  256                                                \\
				14    & conv       & 512     & 3 $\times$ 3 / 1 & 26 $\times$  26 $\times$  256  & 26 $\times$  26 $\times$  512                                                \\
				15    & conv       & 256     & 1 $\times$ 1 / 1 & 26 $\times$  26 $\times$  512  & 26 $\times$  26 $\times$  256                                                \\
				16    & conv       & 512     & 3 $\times$ 3 / 1 & 26 $\times$  26 $\times$  256  & 26 $\times$  26 $\times$  512                                                \\
				17    & max        &         & 2 $\times$ 2 / 2 & 26 $\times$  26 $\times$  512  & 13 $\times$  13 $\times$  512                                                \\
				18    & conv       & 1024    & 3 $\times$ 3 / 1 & 13 $\times$  13 $\times$  512  & 13 $\times$  13 $\times$ 1024                                                \\
				19    & conv       & 512     & 1 $\times$ 1 / 1 & 13 $\times$  13 $\times$ 1024  & 13 $\times$  13 $\times$  512                                                \\
				20    & conv       & 1024    & 3 $\times$ 3 / 1 & 13 $\times$  13 $\times$  512  & 13 $\times$  13 $\times$ 1024                                                \\
				21    & conv       & 512     & 1 $\times$ 1 / 1 & 13 $\times$  13 $\times$ 1024  & 13 $\times$  13 $\times$  512                                                \\
				22    & conv       & 1024    & 3 $\times$ 3 / 1 & 13 $\times$  13 $\times$  512  & 13 $\times$  13 $\times$ 1024                                                \\
				23    & conv       & 1024    & 3 $\times$ 3 / 1 & 13 $\times$  13 $\times$ 1024  & 13 $\times$  13 $\times$ 1024                                                \\
				24    & conv       & 1024    & 3 $\times$ 3 / 1 & 13 $\times$  13 $\times$ 1024  & 13 $\times$  13 $\times$ 1024                                                \\
				25    & route      & 16      &                  &                                &                                                                              \\
				26    & conv       & 64      & 1 $\times$ 1 / 1 & 26 $\times$  26 $\times$ 512   & 26 $\times$  26 $\times$  64                                                 \\
				27    & reorg      &         & / 2              & 26 $\times$  26 $\times$  64   & 13 $\times$  13 $\times$ 256                                                 \\
				28    & route      & 27 24   &                  &                                &                                                                              \\
				29    & conv       & 1024    & 3 $\times$ 3 / 1 & 13 $\times$  13 $\times$ 1280  & 13 $\times$  13 $\times$ 1024                                                \\
				30    & conv       & 720     & 1 $\times$ 1 / 1 & 13 $\times$  13 $\times$ 1024  & 13 $\times$  13 $\times$ 10 $\cdot$ (3$\times$ $N_c$+1+$N_a$+$N_o$)          \\
				31    & prediction &         &                  &                                & 13 $\times$  13 $\times$ 5 $\times$ 2 $\times$ (3$\times N_c$+1+$N_a$+$N_o$) \\
				\hline
			\end{tabular}}
	\end{center}
	\vspace{-5mm}
	\caption{\small Network architecture}
	\label{tab:networkarchitecture}
	\vspace{-5mm}
\end{table}

\vspace{-4mm}

\begin{figure}[h]
	\begin{center}
		{\includegraphics[width=0.8\columnwidth]{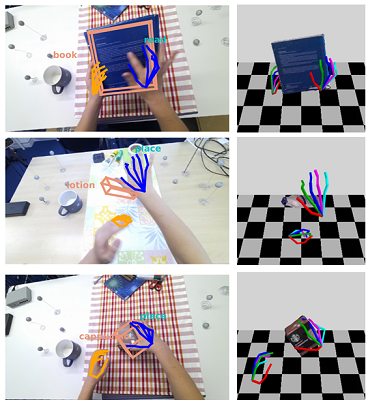}}
		\vspace{-5mm}
	\end{center}
	\caption{\small Some failure cases of our pose prediction method, due to motion blur, reflection and occlusion.}
	\label{fig:failure_cases}
\end{figure}

\paragraph{Implementation details for interaction recognition.} As shown by Eq. 9 in the main paper, the outputs from two $1 \times 1$ convolutional layers, ${\bf W_{\theta}}$ and ${\bf W_{\phi}}$, are multiplied to form a data dependent adjacency matrix, $\bf S_{j}$. This is followed by a softmax layer to normalize the elements in the matrix.

The dimensionality of the input to our overall TA-GCN network is $3\times200\times51$ (for $C \times T \times N$). We use 21 keypoints from left \& right hand as shown in Fig.~\ref{fig:obj_meshes}(b) and 9 keypoints from objects (8 corners and 1 center point of a 3D bounding box). This, in total, results in 51 keypoints fed as input to the network. For inputs larger than 200 frames, we randomly sample 200 frames. For inputs smaller than 200 frames, we pad the data by looping the clip.

Each TA-GCN block takes a $C \times T\times N$ input which is fed into 2D convolutional layer following batch normalization and a ReLu layer. Another batch normalization layer and a dropout layer are placed after the 2D convolutional layer. A skip connection is added to each TA-GCN block to learn more stable features, similarly with  2s-AGCN~\cite{shi2019two}. We set the size of the vertex neighborhood defined by the convolutional kernel as 2. The convolution for the temporal dimension is the same as ST-GCN~\cite{yan2018spatial}.

To build our TA-GCN, we stack 10 TA-GCN blocks. The consecutive numbers of output channels for TA-GCN blocks are 64, 64, 64, 64, 128, 128, 128, 256, 256, and 256. A fully connected layer following average pooling is used as the last layer to predict the class of action labels.
We train the network using SGD with a momentum of 0.9. We set the dropout rate as 0.5 and the batch size as 16. The learning rate starts from 0.005 and is divided by 10 at the 150\textsuperscript{th}, 200\textsuperscript{th} and 250\textsuperscript{th} epoch.

\begin{figure}[t]
	\vspace{-3mm}
	\centering
	\scalebox{1.1}{
		\begin{tabular}{c}
			\hspace{-5mm}\includegraphics[width=\columnwidth]{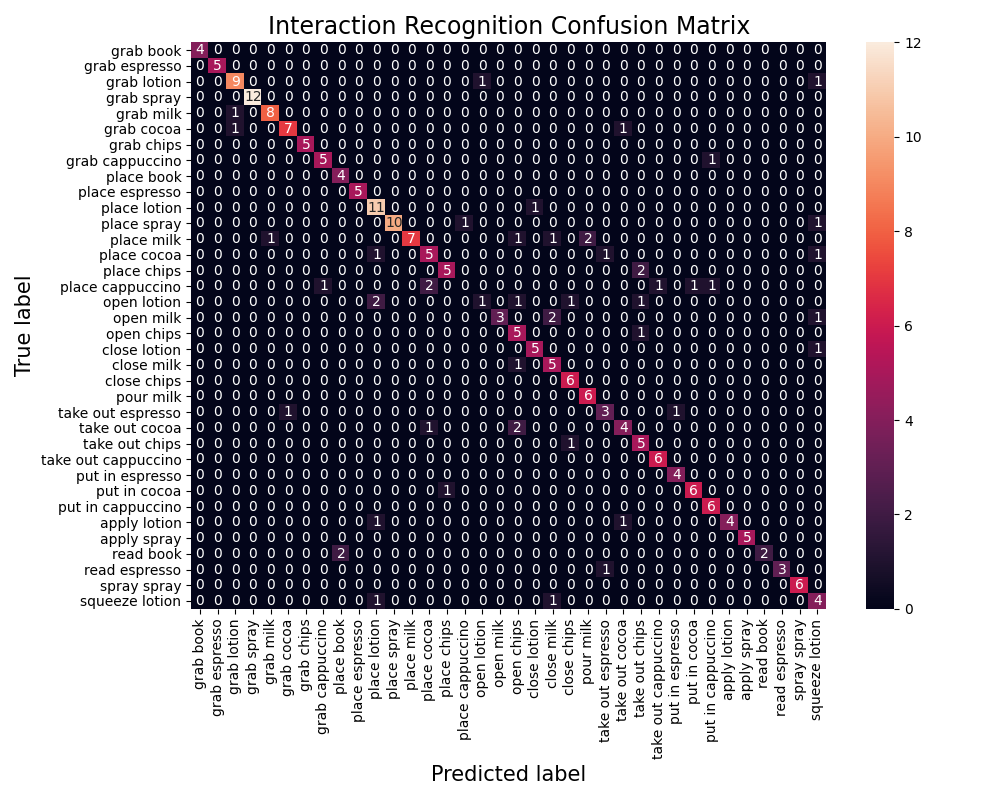}
		\end{tabular}}\vspace{-5mm}
	\caption{\small Confusion matrix for interaction recognition .}
	\label{fig:action_confusion_matrix}
	\vspace{-3mm}
\end{figure}

\begin{table}
	\begin{center}
		\scalebox{0.82}{
			\begin{tabular}{lcc}
				\hline
				                  & Avg. val error (mm) & Avg. test error (mm) \\
				\hline
				Left hand joints  & 22.05               & 41.45                \\
				Right hand joints & 30.12               & 37.21                \\
				Object vertices   & 36.20               & 47.90                \\
				\hline
			\end{tabular}
		}
	\end{center}
	\vspace{-6mm}
	\caption{\small Average errors (in mm) for hand \& object estimates using our pose prediction approach.}
	\label{table:hand_object_average_errors}
	\vspace{-5mm}
\end{table}

\vspace{-4mm}
\paragraph{Evaluation metrics.} In our paper, we use the percentage of correctly estimated poses to assess the accuracy of pose estimation.
Specifically, for hand pose estimation, we use the 3D PCK metric as in~\cite{tekin2019h+} and consider a pose estimate to be correct when the mean distance between  the  predicted  and  ground-truth  joint  positions is less than a certain threshold without a rigid alignment.  When using the percentage of correct poses to evaluate 6D object pose estimation accuracy, we take a pose estimate to be correct if the 2D projection error or the average 3D distance of model vertices is less than a certain threshold (the latter being also referred to as the ADD metric).

\vspace{-4mm}
\paragraph{Confusion matrix for interaction recognition.} We show in Fig.~\ref{fig:action_confusion_matrix} the confusion matrix for interaction recognition. As shown by the strong diagonal of  the  confusion  matrix,  our  model  is  able  to  distinguish between different classes achieving a high accuracy.

\begin{figure*}
	\begin{center}
		\includegraphics[width=1.0\textwidth]{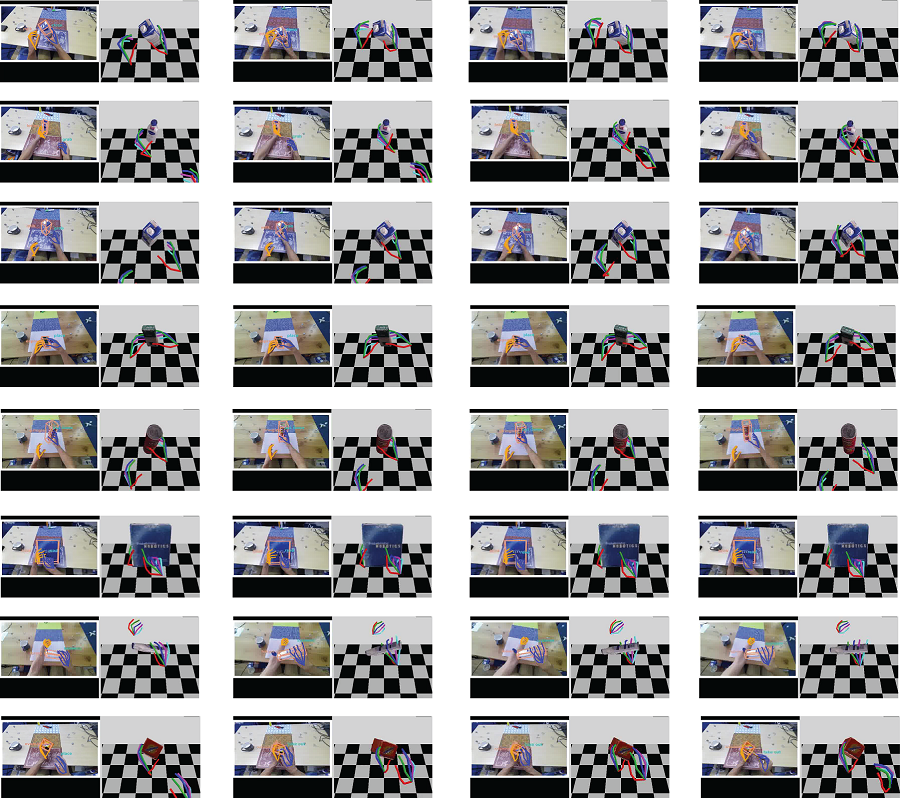}
	\end{center}
	\caption{\small Qualitative results of our method that jointly estimates the poses for two hands \& objects, along with action and object classes.}
	\label{fig:predict_supp}
\end{figure*}

\vspace{-4mm}
\paragraph{Mean errors for hand \& object keypoint prediction.} We further provide average keypoint prediction errors for hands and objects in Euclidean distance in Table~\ref{table:hand_object_average_errors}. Keypoints are selected as 21 joint locations for hands and 21 points on the bounding box (1 center point, 8 corner points, 12 midpoints of the edges) for the object. We demonstrate that, with a low error margin, our method constitutes a strong baseline for joint pose estimation of two hands interacting with objects. We provide further qualitative pose estimation results in Fig.~\ref{fig:predict_supp}.
\\
\\
\\
\\
\\
\\
\\
\\
\\
\\
\\
\\
\\
\\
\begin{figure}[H]
	\begin{center}
		{\includegraphics[width=\columnwidth]{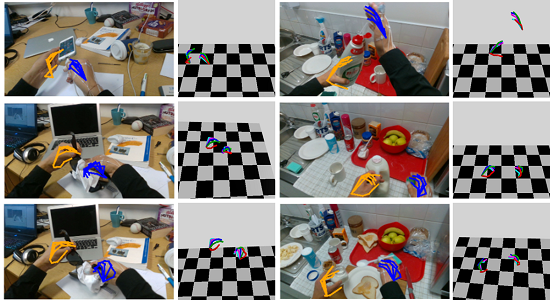}}
		\vspace{-5mm}
	\end{center}
	\caption{\small Inter-dataset examples when we train both hand poses on our dataset with our method and validate on FPHA~\cite{garcia2018first}. Note that the offsets are observed due to different camera parameters across datasets. Polluted images by the magnetic sensors on the FPHA dataset detriment generalization for right hand poses.}
	\label{fig:interdataset_result}
\end{figure}

\end{document}